%% file: aot.tex
\documentclass{article}

\PassOptionsToPackage{numbers}{natbib}
\usepackage[final]{neurips_2025}

\usepackage[utf8]{inputenc} 
\usepackage[T1]{fontenc}    
\usepackage{hyperref}       
\usepackage{url}            
\usepackage{booktabs}       
\usepackage{amsfonts}       
\usepackage{nicefrac}       
\usepackage{microtype}      
\usepackage{xcolor}         
\usepackage{multirow}
\usepackage{graphicx}
\usepackage{amsmath}
\usepackage{subcaption}

\title{Atom of Thoughts for Markov LLM Test-Time Scaling}

\author{%
  Fengwei Teng$^{1,2}$,\quad  Quan Shi$^{3}$,\quad  Zhaoyang Yu$^{2}$,\quad  Jiayi Zhang$^{1,2}$,\quad  \\
  \textbf{Yuyu Luo$^{1}$,\quad Chenglin Wu$^{2\dagger}$,\quad Zhijiang Guo$^{1\dagger}$}\\
  $^{1}$HKUST(GZ),\quad
  $^{2}$DeepWisdom,\quad
  $^{3}$Renmin University of China \\
}

\usepackage{xspace}
\newcommand{\our}{\textsc{AoT}\xspace}
\usepackage{listings}
\usepackage{xcolor}
\usepackage{colortbl}

\definecolor{codegreen}{RGB}{65, 150, 90}
\definecolor{codeblue}{RGB}{66, 91, 201}
\definecolor{codeyellow}{RGB}{201, 154, 66}
\definecolor{codeblack}{RGB}{0, 0, 0}
\definecolor{backcolour}{RGB}{255, 255, 255}
\definecolor{textcolor}{RGB}{0, 0, 0}

\lstdefinestyle{PythonStyle}{
    language=Python,
    basicstyle=\ttfamily\small\color{textcolor},
    backgroundcolor=\color{backcolour},
    commentstyle=\color{codeyellow}\itshape,
    keywordstyle=\color{codeblue}\bfseries,
    stringstyle=\color{codegreen},
    numberstyle=\tiny\color{codeblue},
    breakatwhitespace=false,
    breaklines=true,
    captionpos=b,
    keepspaces=true,
    numbers=none,
    showspaces=false,
    showstringspaces=false,
    showtabs=false,
    tabsize=4,
    frame=leftline,
    framesep=5pt,
    framerule=0.5pt,
    rulecolor=\color{codeblack},
    aboveskip=15pt,
    belowskip=15pt
}

\begin{document}

\maketitle
\begingroup
\renewcommand{\thefootnote}{}
\footnotetext{$^{\dagger}$Corresponding Authors. Contact: steamedbun2002@outlook.com}
\endgroup

\input{sections/0_abstract}
\input{sections/1_intro}

\input{sections/2_related_work}
\input{sections/3_method}

\input{sections/4_experiment}
\input{sections/5_conclusion}

\bibliographystyle{plainnat}
\bibliography{cited}


\input{sections/6_appendix}

\end{document}

%% file: sections/0_abstract.tex
\begin{abstract}
Large Language Models (LLMs) have achieved significant performance gains through test-time scaling methods. However, existing approaches often incur redundant computations due to the accumulation of historical dependency information during inference. To address this challenge, we leverage the memoryless property of Markov processes to minimize reliance on historical context and propose a Markovian reasoning process. This foundational Markov chain structure enables seamless integration with various test-time scaling methods, thereby improving their scaling efficiency. By further scaling up the Markovian reasoning chain through integration with techniques such as tree search and reflective refinement, we uncover an emergent atomic reasoning structure, where reasoning trajectories are decomposed into a series of self-contained, low-complexity atomic units. We name this design Atom of Thoughts (\our). Extensive experiments demonstrate that \our consistently outperforms existing baselines as computational budgets increase. Importantly, \our integrates seamlessly with existing reasoning frameworks and different LLMs (both reasoning and non-reasoning), facilitating scalable, high-performance inference.We submit our code alongside this paper and will make it publicly available to facilitate reproducibility and future research.

\end{abstract}

%% file: sections/1_intro.tex
\section{Introduction}
Large Language Models (LLMs) exhibit remarkable scaling behavior: as model parameters and training data increase, their performance improves predictably across a wide range of tasks~\cite{Kaplan2020scaling, li2025system}. Recently, test-time scaling methods have emerged to push the performance boundary further by increasing computational resources during inference. These range from basic Chain-of-Thought (CoT) prompting that extends reasoning chains~\cite{Wei2022cot}, to more structured approaches like Tree-of-Thought (ToT)~\citep{Yao2023tot} and Graph-of-Thought (GoT)~\citep{Besta2024got} that organize multiple LLM invocations for exploring solution spaces, and recent reasoning models such as OpenAI O1~\citep{O12024} and DeepSeek R1~\citep{deepseekR1Model} that enhance LLMs' long-chain reasoning ability through post-training~\cite{Snell2024ScalingLT, muennighoff2025s1, hou2025advancing}.

\begin{figure*}[t!]
    \centering
    \begin{subfigure}[t]{0.39\textwidth}
        \centering
        \includegraphics[width=\textwidth]{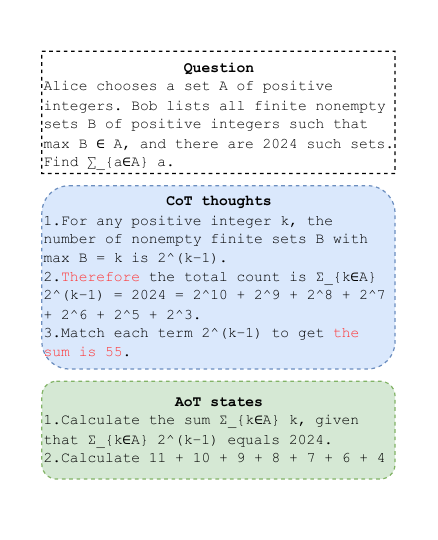}
        \caption{}
        \label{fig:nodes}
    \end{subfigure}
    \hfill
    \begin{subfigure}[t]{0.6\textwidth}
        \centering
        \includegraphics[width=\textwidth]{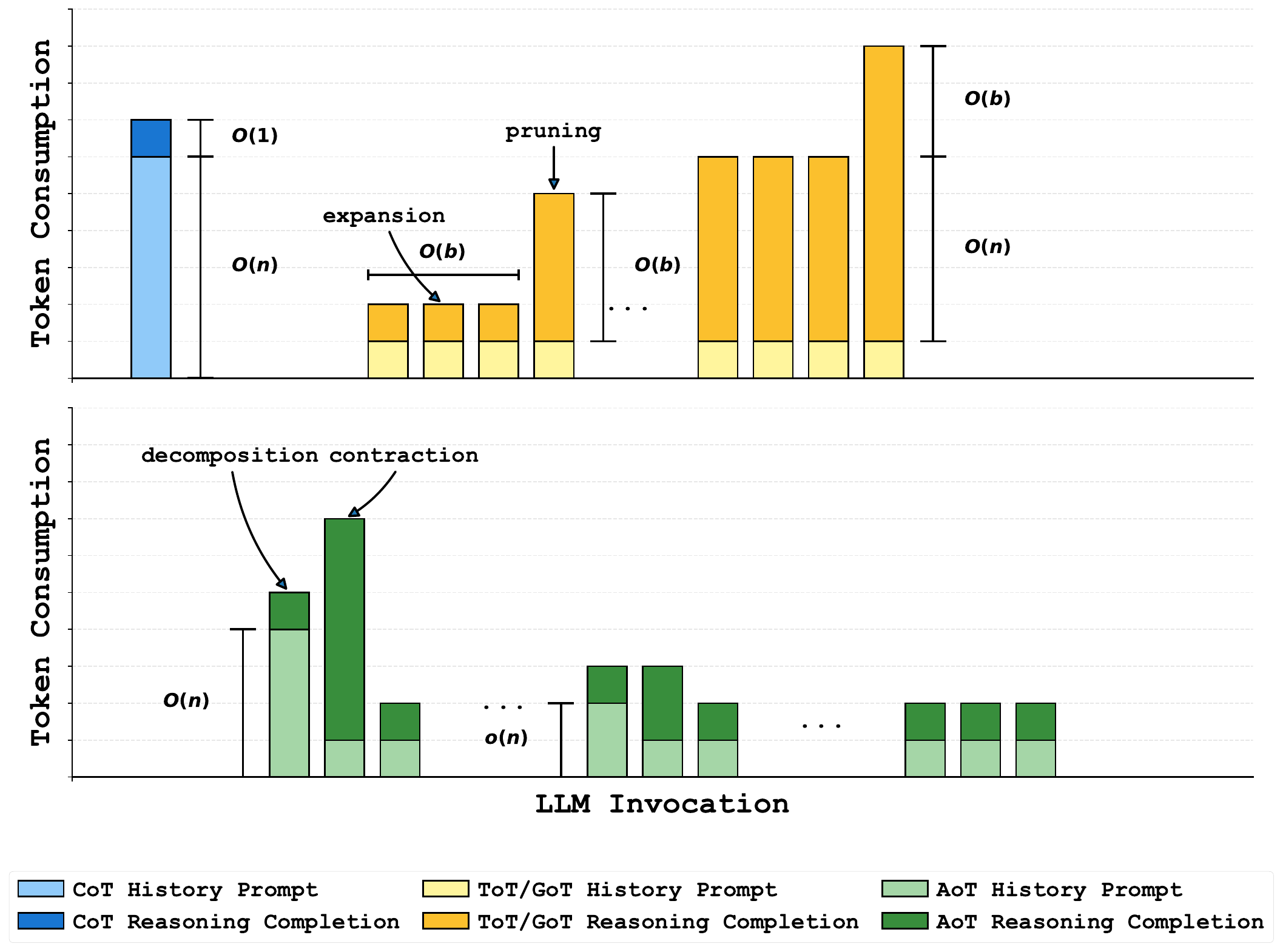}
        \caption{}
        \label{fig:difference}
    \end{subfigure}
    \caption{
    \textbf{Token Allocation Comparison in Reasoning Frameworks.}
    \textit{Figure (a)} demonstrates the differences between thoughts and states, where the red-highlighted text in thoughts reflects dependencies on historical information, whereas states maintain answer-equivalence with the initial problem while progressively reducing execution complexity. 
    \textit{Figure (b)} illustrates differences in the number of prompt tokens and completion tokens for CoT, ToT, GoT, and the state-based \our. For simplicity, we assume each thought consists of the same number of tokens, with an average of \(O(n)\) thoughts required to express a solution. While ToT maintains \( b \) branches, resulting in a fixed number of \( b \) invocations per expansion stage, GoT's settings can be flexibly adjusted depending on the scenario and are thus denoted as \( O(b) \). 
    }
    \label{fig:combined}
\end{figure*}

However, current framework-based test-time scaling methods typically rely heavily on retaining extensive historical information. Even the simplest CoT must preserve the entire reasoning trajectory to generate each subsequent step~\cite{Wei2022cot, Zhang2024autocot}. Tree-based methods maintain ancestor and sibling relations for branching decisions~\cite{Yao2023tot, Zhou2024lats, Ding2023xot}, while Graph-based methods introduce even more complex dependencies through arbitrary node connections~\cite{Besta2024got, Zhang2024dot}. Figure~\ref{fig:difference} analyzes these representative structures and abstract the complexity of historical information and reasoning completion token involved at each LLM invocation.

To decouple the current problem’s reasoning from processing historical information and thus minimize their mutual interference during test-time computation, we aim to generalize Markov chain–style structures to general-purpose reasoning. By exploiting the \textbf{memoryless property} of Markov processes, we design the Markovian reasoning process, where each state encapsulates a self-contained problem, thereby significantly reducing historical dependencies. The reasoning process is expressed as a sequence of states with progressively reduced test-time complexity, rather than an accumulation of historical thoughts like CoT, as illustrated in Figure~\ref{fig:nodes}. To ensure steady progress, we introduce a two-phase state transition mechanism: the decomposition stage converts the current state into a Directed Acyclic Graph (DAG)-based reasoning path, and the contraction stage uses its structure to reduce dependencies and generate the next state.

This fundamental structure with memoryless property distinguish our approach from many CoT-based methods. Thus, our method can be seamlessly integrated with existing test-time scaling methods, enhancing their scaling efficiency. While exploring integrations with tree search and reflective refinement to further scale up the Markovian reasoning chain, we identify an emergent trend towards an atomic reasoning structure (Figure~\ref{fig:integration}), where reasoning trajectories are represented as a series of self-contained, low-complexity atomic problems. To emphasize this characteristic, we name our approach Atom of Thoughts (\our).

Our contributions are summarized as follows:
\begin{itemize}
    \item \textbf{Markovian Reasoning Process.}
    We introduce a general-purpose Markovian reasoning process that achieves high-quality and cost-effective reasoning across various scenarios, including code generation, mathematical reasoning, and multi-step reasoning tasks.
    
    \item \textbf{Scalable Reasoning Structure.} 
    The basic structure design of Markov chain in \our facilitates seamless integration with various test-time scaling methods, significantly enhancing computational efficiency and allowing the combination of different methods' advantages. This scalability ensures more effective utilization of increased computational budgets without the overhead of maintaining extensive historical contexts.
    
    \item \textbf{Atomic Reasoning.} 
    Further leveraging \our's seamless integration capability to enhance itself, by integrating with tree search and reflective refinement to scale up the exploration of the Markovian reasoning chain, we uncover an emergent atomic reasoning structure. In this structure, complex reasoning trajectories are decomposed into a sequence of atomic, self-contained units with low complexity. This atomicization brings about improved reasoning performance and robustness.
\end{itemize}

%% file: sections/2_related_work.tex
\section{Related Work}
\label{sec:related_work}

\subsection{Reasoning Framework}

Drawing inspiration from cognitive behaviors in human reasoning~\cite{Smelser2001Cognitive}—such as step-by-step decomposition~\cite{Wei2022cot, zhou2023least, wang2023planandsolve, hao2024llm}, reflective refinement~\cite{Madaan2023selfrefine, zheng2024stepback, zheng2023progressive, zhan2025evaluating}, and aggregation ensemble~\cite{Wang2023cotsc, jiang2023llmblender, yao2025determine}—various prompting strategies have been developed to enhance the reasoning capabilities of LLMs. These reasoning frameworks typically employ structured representations, including chains, graphs, and trees~\cite{Yao2023tot, Besta2024got, zhang2024tse}, to model the reasoning space efficiently and systematically. Chain-based methods, for instance, decompose complex problems into linear sequences of subproblems~\cite{Wei2022cot, zhou2023least, wang2023planandsolve}, primarily optimizing for stepwise dependency. In contrast, tree- and graph-based formalisms support hierarchical exploration of multiple reasoning paths, allowing for more dynamic adaptation during the problem-solving process~\cite{Yao2023tot, Besta2024got}. These structured approaches have demonstrably improved LLM performance in diverse applications like code generation, question answering, and complex data processing~\cite{Hong2024metagpt, Hong2024data, zhang2025evoflow, zhang2024mobileexperts}, by enabling LLMs to tackle intricate problems with enhanced coherence and interpretability.

While these structured methods significantly expand LLMs' reasoning capabilities, they also inherently accumulate historical dependencies. This accumulation can lead to increased computational costs and potential interference during the inference process. Recent efforts have attempted to mitigate this reliance on historical information by exploring Markovian reasoning processes and atomic reasoning steps, aiming for more memoryless transitions~\cite{Xin2024atomr, Xiang2024AtomThink, Zhou2024selfdiscover, xiang2025can}. However, these approaches often suffer from task-specific design limitations, hindering generalizability and efficient parallelism~\cite{hao2023rap, Yang2024mcot, Zekri2024Large}. In contrast, 
\our introduces a DAG-based approach that decouples partial subproblems into atomic nodes. This decoupling enables independent state transitions without the substantial overhead associated with maintaining historical context. By iteratively decomposing problems into these atomic nodes and then contracting them, our method reduces overall complexity and inherently supports efficient parallel execution, thereby addressing the limitations of traditional chain, tree, and graph-based structures.

\subsection{Test-Time Scaling}
Test-time scaling has emerged as a powerful mechanism to enhance LLM reasoning by extending computational effort during inference. Framework-based approaches augment LLM capabilities through structured reasoning extensions, leveraging cognitive operations and external tool integration to facilitate deeper exploration of solution spaces~\cite{zhang2024aflow, Falcon2024archon, Chen2024more}. These methods introduce reflective reasoning cycles, recursive problem-solving, and dynamic path selection, significantly improving performance on complex reasoning tasks. Despite these advances, existing techniques commonly preserve full historical state information throughout the reasoning process. This can lead to redundant computational overhead and potential conflicts across successive reasoning steps.

Recent work has explored alternative strategies, such as supervised fine-tuning on CoT trajectories, demonstrating improved LLM capacity to maintain coherent, long-term reasoning~\cite{ye2025limo, yeo2025demystifying, yao2025unveiling, song2025r1, song2025r1pp, sun2025simpledeepsearcher}. Reinforcement learning have further pushed these boundaries by enabling models to autonomously extend reasoning chains, potentially unlocking emergent cognitive patterns~\cite{team2025kimi, zeng2025simplerl, deepseekR1Model, yu2025dapo}. However, similar to framework-based methods, these techniques often rely on maintaining expansive historical contexts, which can limit their efficiency and scalability as reasoning paths become extended.

In contrast to these history-dependent methods, our approach adopts a Markovian perspective, modeling the reasoning process as state transitions assisted by a temporary DAG structure. This memoryless design eliminates the need for redundant history tracking, focusing computational resources solely on current state transformations. Furthermore, our proposed two-phase transition mechanism, comprising decomposition and contraction stages, facilitates atomic problem-solving. This enhances computational efficiency while maintaining structural clarity. This structured yet flexible approach not only reduces dependency overhead but also aligns naturally with the principles of test-time scaling, offering seamless integration with existing reasoning frameworks to achieve scalable, high-performance inference.

%% file: sections/3_method.tex
\section{Atomic Reasoning via Markov Process}
\label{sec:method}
\begin{figure*}[t!]
    \centering
    \includegraphics[width=\textwidth]{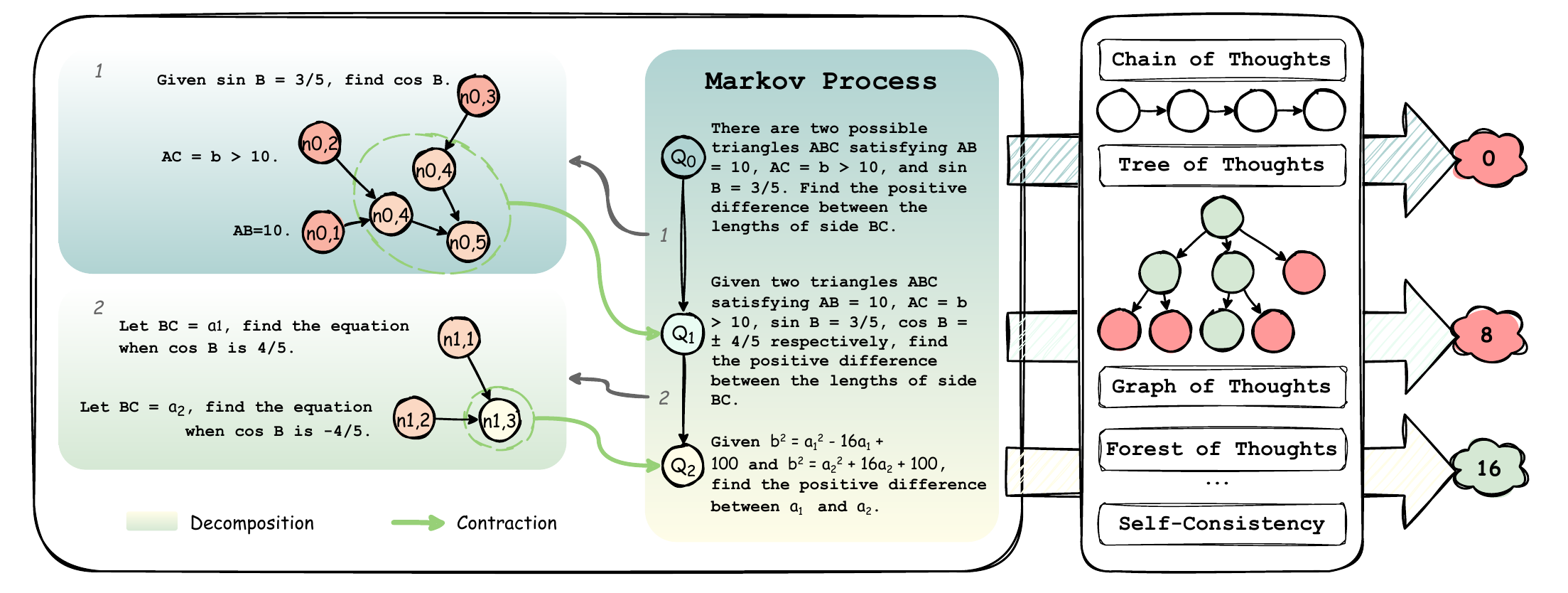} 
    \caption{\textbf{Overview of \our.} 
    The Markov reasoning framework iteratively derives states \( Q_{i+1} \) from predecessors \( Q_{i} \) using DAG decomposition and contraction.
    The left part shows this iterative process, while the right part highlights the integration with existing methods. Any intermediate state \( Q_i \) can act as an entry point \( Q_0 \) for other methods, ensuring flexible composition while preserving answer equivalence to the original question. This allows \our to operate independently or as a preprocessing module to optimize the performance or efficiency of existing approaches.}
    \label{fig:pipeline}
\end{figure*}

In this section, we first formally derive a Markovian reasoning process grounded in a clear probabilistic formulation. We then discuss how this Markovian reasoning structure can be integrated seamlessly with other reasoning methods to further scale up inference time. Finally, we demonstrate how atomic reasoning structures naturally emerge through such scaling-up procedures. The overview of this Markovian reasoning process is illustrated in Figure \ref{fig:pipeline}.

\subsection{Markovian Reasoning Process}

\paragraph{Reasoning Chain.}  
CoT reasoning introduces a sequence of intermediate steps \( T_i \) to solve a problem. This process can be formalized as a probabilistic sampling procedure:
\begin{align}  
A \sim p(A|\mathcal{T}, Q_0) \prod_{i=0}^N p(T_i|\mathcal{T}_{<i}, Q_0)  
\end{align}
where \( A \) is the final answer, and \( \mathcal{T} = \{T_0, T_1, \dots, T_N\} \) is the sequence of thoughts, each conditioned on the previous steps \( \mathcal{T}_{<i} \) and the initial question \( Q_0 \).

An alternative formulation—Least-to-Most~\cite{zhou2023least} prompting—reframes the node of chain as a subquestion \( Q_i \), yielding:
\begin{align}  
A \sim p(A|\mathcal{Q}) \prod_{i=0}^N p(Q_i|\mathcal{Q}_{<i})  
\end{align}

Under the above formulation, the reasoning process is characterized by the accumulation of intermediate thoughts or subquestions in the sequence, leading to a continual increase in historical information.  
However, ideally, if the reasoning chain satisfies the property of a memoryless Markov process—where each state \( S_{i+1} \) depends only on \( S_i \)—we obtain:
\begin{align}
A \sim p(A|S_N) \prod_{i=0}^N p(S_{i+1}|S_i)
\end{align}
where \( S_i \) represents a state in the Markovian reasoning process. In the following paragraph, we will explicitly clarify the semantic content of the Markov state \( S_i \), resulting in a more specific and practical representation.

\paragraph{Markov State.}
In practice, real-world problems rarely satisfy the strict Markov assumption directly. To establish a meaningful Markovian formulation, we reuse the subquestion symbol \( Q_i \) to represent the Markov states \( S_i \), initialized by the original question \( Q_0 \). Since the final answer \( A \) must be derivable from the final state \( Q_{-1} \), it follows naturally that \( Q_{-1} \) is answer-equivalent to \( Q_0 \). Thus, an essential invariant emerges: each intermediate subquestion \( Q_i \) must preserve answer-equivalence with the original question. To ensure meaningful Markov state transitions, we further impose that the sequence of subquestions \(\{Q_0, Q_1, \dots, Q_N\}\) monotonically reduces in complexity, guaranteeing genuine reasoning progress at each transition.

\paragraph{Two-phase Transition}  
However, state transitions aiming at test-time reduction remain challenging for LLMs, especially without task-specific training. This difficulty arises primarily from the complex historical dependencies within reasoning trajectories. To address this issue, we propose a two-phase transition mechanism that first explicitly decomposes the current state \( Q_i \) to capture the internal dependencies before contracting them into the next state.

In the decomposition phase, we introduce a DAG scaffold $\mathcal{G}_i$ to explicitly represent the dependency structure among reasoning steps within each intermediate question $Q_i$. This temporary structure is later discarded to eliminate historical dependencies, enabling the Markovian transition. Formally, the DAG is defined as:
\begin{align}
\mathcal{G}_i = (\mathcal{N}, E), \quad E \subseteq \{(N_j, N_k) \mid j < k\}
\end{align}
where nodes $N_k$ represent individual thoughts or subquestions, and edges $(N_j, N_k)$ indicate that node $N_j$ provides necessary information for node $N_k$.

In the subsequent contraction phase, we transform the temporary DAG structure $\mathcal{G}_i$ into the next Markov state $Q_{i+1}$. Specifically, nodes without incoming edges in $\mathcal{G}_i$ are independent and can be safely discarded, whereas the remaining dependent nodes are reformulated into an answer-equivalent independent question $Q_{i+1}$. Formally, the overall Markovian transition process can be expressed as:
\begin{align}
A \sim p(A|Q_N)\prod_{i=0}^{N} p(Q_{i+1}|\mathcal{G}_i)\, p(\mathcal{G}_i|Q_i).
\end{align}
A detailed step-by-step example demonstrating the complete decomposition-contraction process is provided in Appendix~\ref{appendix:detailed_examples}.

\subsection{Emerged Atomic Reasoning}
The Markovian reasoning process provides a fundamental, low-level structural prior for inference. In this subsection, we discuss the design of a termination mechanism to counteract the potential fragility introduced by strict memorylessness, thereby constructing a stable reasoning framework. Moreover, we describe how this Markovian reasoning structure can be combined with additional methods—particularly through structured exploration via tree search and reflective verification—to further scale up test-time reasoning. This combined approach reveals the emergence of a stable, indivisible reasoning structure, termed atomic reasoning.

\paragraph{Termination Strategy.}  
Unlike CoT-based approaches, which can recover from early errors by leveraging accumulated context, our Markov chain lacks such a fallback due to its memoryless nature. This amplifies the risk of propagating low-quality transitions—if an intermediate question \( Q_{i+1} \) diverges semantically from the original task, subsequent reasoning becomes meaningless. 

To address this, we introduce a quality-aware termination strategy. After each transition \( Q_i \rightarrow Q_{i+1} \), an LLM-as-a-judge selects the best answer to the original question \( Q_0 \) from the triplet \( \{\text{solve}(Q_i), \text{solve}(\mathcal{G}_i), \text{solve}(Q_{i+1})\} \). Crucially, this mechanism implicitly enforces answer equivalence: if \( Q_{i+1} \) fails to preserve answer equivalence with \( Q_0 \), then \( \text{solve}(Q_{i+1}) \) will not provide a valid answer for \( Q_0 \) and thus cannot be selected by the judge. This selection-based filtering naturally ensures that only semantically stable transformations maintaining answer equivalence are retained. If \( Q_{i+1} \) is not selected, the process terminates and returns the best candidate among the three. Detailed quality metrics demonstrating the effectiveness of this mechanism are provided in Appendix~\ref{appendix:dag_quality}.

\paragraph{Modular Integration.}
Since each Markov state is constrained to be an equivalently transformed representation of the original question, the reasoning process forms a semantically aligned and fully self-contained sequence of problem representations. This property enables modular reasoning without compromising the integrity of the overall task. In practice, each state within the chain can be independently routed to specialized solvers, subjected to verification procedures, or further embedded into structured reasoning frameworks—such as tree-based or graph-based inference. The introduction of the Markov reasoning process thus does not merely offer an alternative to previous reasoning chain methods, but rather defines a structural foundation upon which diverse test-time reasoning strategies can be constructed.

\paragraph{Atomic Structure.}
Although the termination strategy ensures robustness, it also restricts the emergence of deeper reasoning chains. To explore the full potential of the Markov process, we sample and extend trajectories, combining tree search and reflection mechanisms. These structured explorations reveal a statistically supported phenomenon: deeper reasoning states tend to converge into irreducible forms, maintaining a stable and relatively low reasoning token count, from which the original problem's answer can be directly inferred with high execution stability. We refer to these stable forms as atomic structures: indivisible and self-contained representations that require no further decomposition. Importantly, atomicity is not imposed a priori, but emerges naturally as a property discovered throughout the reasoning process. This convergence toward atomic units represents a logical endpoint where problems become sufficiently simple that further decomposition is neither necessary nor beneficial. Notably, this convergence point is jointly determined by both the intrinsic complexity of the problem and the reasoning capabilities of the underlying model—different problems may converge at different depths, and the same problem may exhibit different atomic granularities when solved by models with varying capacities.

%% file: sections/4_experiment.tex
\section{Experiments}
\label{sec:experiment}

\input{tables/main_result}

Our experiments aim at two primary objectives. First, we conduct main experiments across a variety of datasets spanning mathematics, code generation, and multi-hop question answering to demonstrate the cost-efficiency advantages of \our as a general-purpose reasoning framework. Second, leveraging the flexibility provided by the basic Markov chain structure in our approach, we design integration experiments at various granularities. These experiments explore the utilization of \our as a plug-in component to enhance cost-efficiency in other reasoning frameworks and investigate scaling effects in integration with classical methods like tree search and verification-based reflection, analyzing emergent reasoning phenomena.

\subsection{Experimental Setup}

\paragraph{Benchmarks and Metrics.} We evaluate \our across representative benchmarks covering mathematical reasoning (MATH~\cite{MATH2021}, GSM8K~\cite{gsm8k2021}, AIME\footnote{https://huggingface.co/datasets/Maxwell-Jia/AIME\_2024}), code generation (MBPP~\cite{mbpp2021Austin}, LiveCodeBench~\cite{jainlivecodebench}), and multi-hop question answering tasks (HotpotQA~\cite{HotpotQA2018}, MuSiQue~\cite{Musique2022}, and 2WikiMultiHopQA~\cite{2WikiMultiHopQA} preprocessed by LongBench~\cite{longbench2024}), see Appendix F.3 for details. Following previous work~\citep{zhang2024aflow,Bi2024fot}, we report pass rates for mathematical and coding benchmarks, and F1 scores for multi-hop QA tasks.

\paragraph{Settings.}
All prompt templates used in Markov reasoning process for experiments are fully described in Appendix~\ref{sec:prompt}. Key hyperparameters, including model temperature and Markov chain length, are detailed and discussed in Appendix~\ref{sec:hyperparams}. We set the default temperature to 1.0 and the maximum Markov chain length to 3 for the main experiments to balance performance and efficiency while enabling scaling curves. Due to \our’s design and termination mechanism, longer chain lengths increase the performance ceiling without linearly increasing costs.

\begin{figure*}[t!]     
    \centering     
    \includegraphics[width=\textwidth]{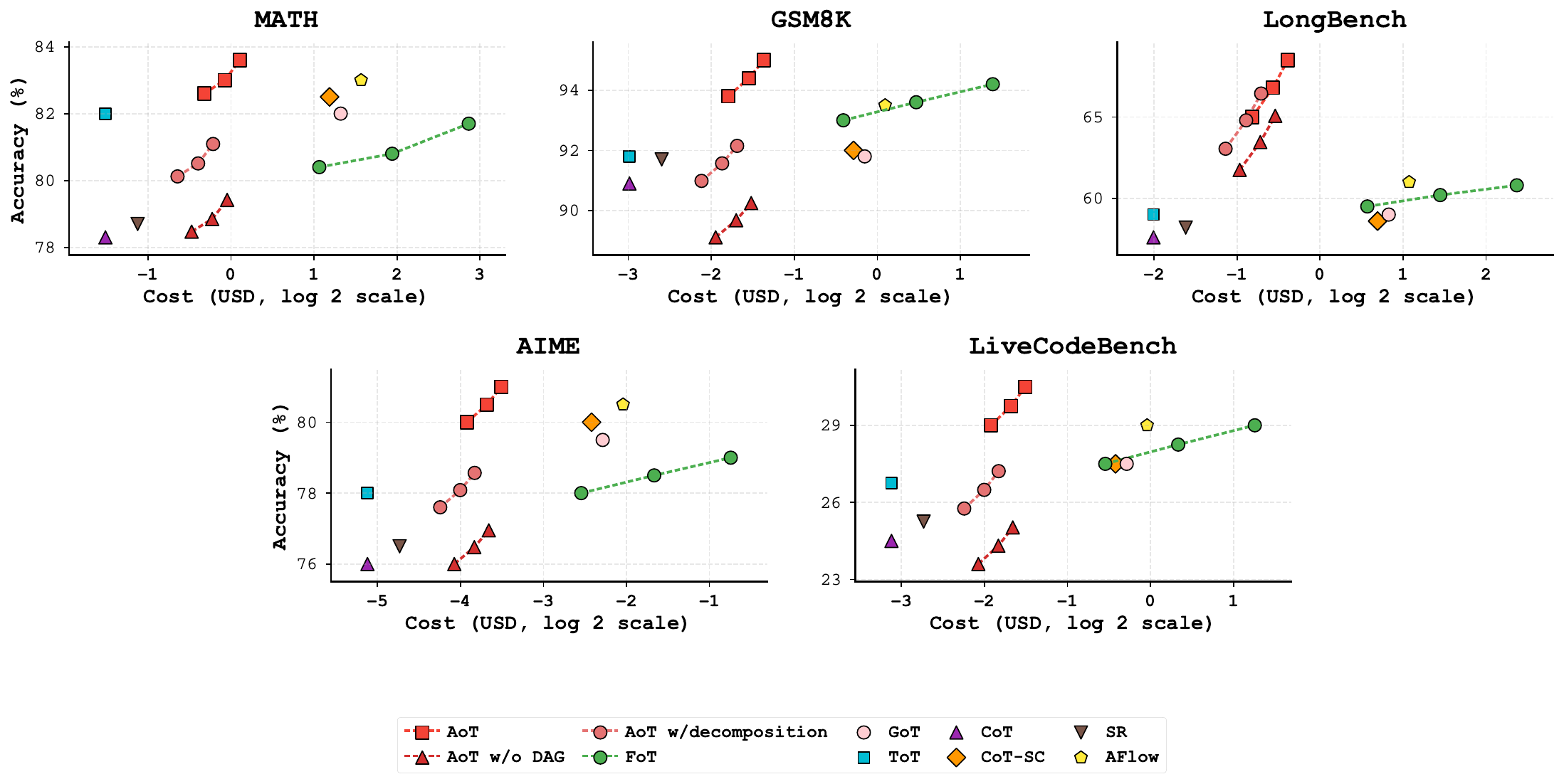}      
    \caption{A comparison of performance and cost of various methods and ablation methods on the dataset, with GPT-4o-mini as the backbone. Each node in the curves represents an AoT (or ablation variants) iteration result, where increasing token consumption indicates deeper iterations. Due to relatively poor AR performance leading to scattered data points, AR data points are excluded.}
    \label{fig:main} 
\end{figure*}
\paragraph{Backbones and Baselines.}
\our is designed to be compatible with various LLM backbones. To demonstrate its effectiveness, we employed two categories of LLMs. The first category comprises non-reasoning LLMs, specifically GPT-4o-mini~\citep{GPT4omini} and DeepSeek-V3~\citep{DeepSeekV3}. The second category includes reasoning-capable LLMs such as O3-mini~\citep{O3miniModel} and DeepSeek R1~\citep{deepseekR1Model}. Specifically, we use non-reasoning models to evaluate performance on MATH, GSM8K, and MBPP, and reasoning-capable models to evaluate performance on more challenging tasks such as AIME and LiveCodeBench. Additionally, since multi-hop QA is not a primary focus for reasoning-capable models, both categories of models are evaluated on LongBench for comprehensive comparison.

For comparison, we evaluated \our against a diverse set of baseline methods, broadly categorized by their interaction pattern with the LLM: single-call or multi-call invocations. Single-call approaches include well-known techniques like Chain-of-Thought (CoT)~\citep{Wei2022cot} and Chain-of-Draft (CoD)~\citep{CoD2025}. Multi-call methods represent more complex workflows, such as CoT with Self-Consistency (CoT-SC)~\citep{Wang2023cotsc}, Self-Refine (SR)~\citep{Madaan2023selfrefine}, Analogical Prompting (AP)~\citep{Yasunaga2024AP}, Forest-of-Thought (FoT)~\citep{Bi2024fot}, and the agentic framework AFlow~\citep{zhang2024aflow}. Further details are provided in Appendix~\ref{appendix:baseline_implementation}.
\subsection{Main Results}

Table~\ref{tab:performance} presents the main experimental results. 
Across both Non-Reasoning and Reasoning LLMs, \our consistently demonstrates strong performance. For Non-Reasoning LLMs such as GPT-4o-mini and DeepSeek-V3, \our achieves the highest scores on benchmarks like MATH, GSM8K, MBPP, and LongBench, often surpassing all other compared methods. For instance, with GPT-4o-mini, \our scores 83.6 on MATH, 95.0 on GSM8K, 75.2 on MBPP, and 68.5 on LongBench, which are the top performances. Similarly, DeepSeek-V3 with \our leads with scores on all benchmarks.

In the Reasoning LLMs section, featuring O3-mini and DeepSeek-R1, AoT continues to exhibit competitive and often leading performance. For O3-mini, AoT achieves the highest scores on AIME (83.0), LiveCodeBench (32.2), and LongBench (65.3). With DeepSeek-R1, AoT again leads on all tasks. Overall, \our consistently achieves state-of-the-art or highly competitive results across a diverse set of models and benchmarks, demonstrating its effectiveness.

Figure~\ref{fig:main} further demonstrates that performance improves progressively with additional reasoning iterations. This highlights the effectiveness of our proposed termination strategy: by mitigating error propagation from memoryless Markovian transitions, it preserves the desirable test-time scaling property—performance does not degrade as more computational resources are allocated.

\subsection{Ablation Study}
We conduct ablation studies to examine the impact of core components in our framework. Specifically, we evaluate two variants: (1) Without Decomposition, where the model directly contracts reasoning trajectories from the initial question without constructing a DAG; and (2) Without DAG-guided Contraction, where decomposition still occurs, but the contraction step does not rely on any structural guidance. In this setting, only the first naturally independent subproblem is separated out. Figure~\ref{fig:main} shows that both ablations significantly degrade performance, with the second variant causing a more severe drop. This suggests that partial or superficial structural cues can be more harmful than providing none at all. These results underscore the importance of explicitly modeling fine-grained dependencies in reasoning trajectories, showing that faithful structural representations meaningfully enhance reasoning effectiveness and precision. Comprehensive quality metrics for the DAG generation process, including answer equivalence maintenance rates (>99\% across all datasets) and complexity reduction rates (74-82\%), are provided in Appendix~\ref{appendix:dag_quality}.

\subsection{Scaling Up Analysis}
\label{sec:integration_analysis}
In this section, we further explore the scalability of \our by integrating it with existing reasoning frameworks, leveraging its flexible, modular design. We begin our analysis by using individual Markov states as integration points—a lightweight and straightforward approach where intermediate states processed by \our serve as optimized entry points for other reasoning methods. Our experiments reveal substantial efficiency improvements at test-time, which encourages us to examine larger, more structured integration granularities to fully capitalize on the structural strengths of our framework. Notably, as we progressively extend the Markov chain during scaling analysis, we observe a consistent reduction in the number of tokens required for reasoning in the final states. Through detailed analysis, we identify emerging atomic characteristics in the reasoning trajectories, motivating us to design further scaling-up experiments based on this property.

\begin{figure*}[t!]
    \centering
    \includegraphics[width=\textwidth]{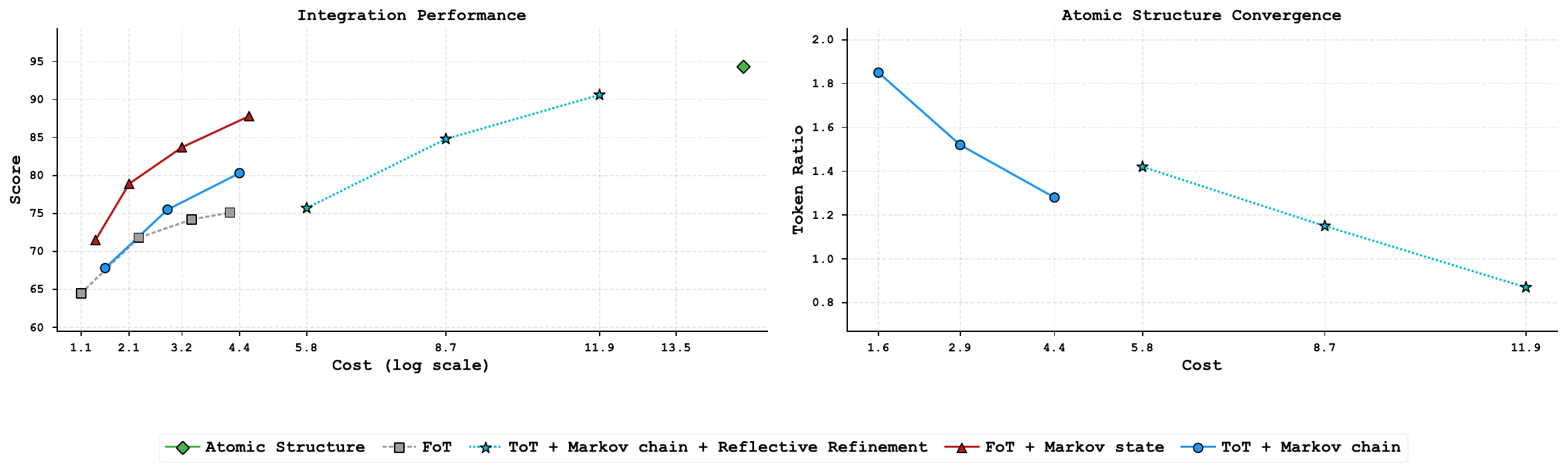} 
    \caption{The process involves gradually enhancing integration for scaling up at test time. ToT uses three branches, while FoT employs two, four, and eight trees, respectively.}
    \label{fig:integration}
\end{figure*}

\paragraph{State Integration.} The Markov states \( Q_i \) generated by \our represent simplified, yet answer-equivalent reformulations of the original questions, making them ideal entry points for external methods. Indeed, \our itself demonstrates such modular integration potential, employing basic CoT-style prompting to solve each intermediate state. To experimentally validate the effectiveness of these intermediate states, we investigate whether initiating reasoning using optimized intermediate states \( Q_1 \) can enhance both accuracy and computational efficiency in external frameworks. The results, illustrated in Figure~\ref{fig:integration}, confirm that starting reasoning from these optimized intermediate states notably improves performance while simultaneously reducing computational costs, as demonstrated in the integration with frameworks such as FoT.

\paragraph{Tree Searching.} Beyond single-state integration, the full Markov sequence \( \mathcal{Q} \) generated by \our can provide a structured scaffold for more complex reasoning frameworks, effectively replacing traditional CoT-based structures. In conventional CoT-based ToT, the inherent randomness of LLM-based sampling can lead to inconsistencies in reasoning chain lengths, causing nodes at the same depth to represent varying stages of reasoning progress. This inconsistency complicates node comparison and diminishes pruning effectiveness. In contrast, the Markov chains constructed by \our ensure answer equivalence between each intermediate node and the original question, thereby guaranteeing fundamental comparability across nodes at the same depth. This structural consistency significantly enhances the gains from scaling through parallel sampling at test-time.

\paragraph{Reflective Refinement.} Termination strategy in \our provides a safeguard for the quality of single-pass Markov reasoning. When a transition yields a low-quality intermediate state, early termination allows the system to avoid wasting computation on unpromising paths. However, this conservative mechanism may also limit further exploration. To address this, we augment our method with verification-based reflection, where transitions $Q_i \rightarrow Q_{i+1}$ are evaluated by an LLM-as-a-judge to assess whether the newly generated state exhibits a significant degradation in test-time performance. If such degradation is detected, the system triggers a reflective refinement step, encouraging deeper and more meaningful reasoning rather than trivial reformulations. This reflective verification substantially improves comparability between nodes at the same depth, increases the effective exploration space, and further amplifies the benefits of structural scaling. When combining all three integration strategies (ToT + Markov chain + Reflective Refinement), we observe significant performance gains: for instance, on MATH, this full integration achieves 84.9\% accuracy compared to ToT's 82.0\%, and on AIME, it reaches 81.2\% versus ToT's 78.0\%, demonstrating the compounding benefits of our modular design.

\paragraph{Atomic Struture.} Due to the inherent scalability of the \our architecture, deeper Markov chains---enabled by both tree search and verification-based reflection---exhibit stronger test-time performance and require fewer reasoning tokens in the final state. Statistical analysis reveals that the token count of final reasoning steps gradually approaches that of a minimal DAG representation comprising all independent subproblems generated during transitions. This suggests a natural convergence toward atomic states---questions that are semantically represent indivisible reasoning units. We refer to this phenomenon as atomic reasoning, where the entire reasoning trajectory is composed of such minimal, non-decomposable elements. To further validate this insight, we conduct an additional experiment where we isolate and re-execute these highly atomic reasoning paths independently. While this incurs significantly higher computational cost, the results exhibit stable scaling trends, highlighting the structural advantages of \our with high budget.

%% file: tables/main_result.tex
\begin{table}[t!]
    \centering
    \caption{Performance Comparison.}
    \label{tab:performance}
    \scriptsize 
    \begin{tabular}{llcccccccccc}
        \toprule
        \textbf{Model} & \textbf{Benchmark} & \textbf{CoT} & \textbf{CoT-SC} & \textbf{SR} & \textbf{AR} & \textbf{AFlow} & \textbf{ToT} & \textbf{GoT} & \textbf{FoT} & \textbf{AoT} \\
        \midrule
        \multicolumn{11}{c}{\textbf{Non-Reasoning LLMs}} \\
        \midrule
        \multirow{4}{*}{GPT-4o-mini} 	
        & MATH      & 78.3 & 81.8 & 78.7 & 65.4 & \underline{83.0} & 82.0 & 82.3 & 82.6 & \textbf{83.6} \\
        & GSM8K     & 90.9 & 92.0 & 91.7 & 87.2 & 93.5 & 91.8 & 92.1 & \underline{94.2} & \textbf{95.0} \\
        & MBPP      & 72.4 & 73.2 & 72.8 & 70.1 & 74.0 & 73.5 & 73.7 & \underline{74.8} & \textbf{75.2} \\
        & LongBench & 57.6 & 58.6 & 58.2 & 52.9 & \underline{61.0} & 59.0 & 59.2 & 60.8 & \textbf{68.5} \\
        \midrule
        \multirow{4}{*}{DeepSeek-V3} 
        & MATH      & 94.4 & 95.2 & 94.8 & 90.1 & \underline{96.1} & 95.0 & 95.3 & 95.6 & \textbf{96.5} \\
        & GSM8K     & 96.2 & 97.0 & 96.8 & 92.5 & \underline{97.8} & 96.5 & 96.8 & 97.5 & \textbf{98.2} \\
        & MBPP      & 75.7 & 76.5 & 76.0 & 73.2 & 77.3 & 76.8 & 77.0 & \underline{78.2} & \textbf{79.6} \\
        & LongBench & 58.8 & 60.1 & 59.5 & 55.3 & \underline{63.5} & 61.2 & 61.5 & 63.3 & \textbf{71.0} \\
        \midrule
        \multicolumn{11}{c}{\textbf{Reasoning LLMs}} \\
        \midrule
        \multirow{3}{*}{O3-mini} 
            & AIME      & 79.6 & 81.0 & 80.2 & 76.0 & \underline{82.5} & 81.2 & 81.5 & 81.8 & \textbf{83.0} \\
            & LiveCodeBench & 23.6 & 25.0 & 24.2 & 20.0 & 26.5 & 25.2 & 25.5 & \underline{27.8} & \textbf{32.2} \\
            & LongBench & 56.3 & 57.5 & 56.8 & 52.0 & 58.0 & 56.5 & 56.8 & \underline{57.9} & \textbf{65.3} \\
        \midrule
        \multirow{3}{*}{DeepSeek-R1} 
            & AIME      & 78.3 & 79.7 & 78.9 & 74.7 & \underline{81.2} & 79.9 & 80.2 & 80.5 & \textbf{81.7} \\
            & LiveCodeBench & 24.5 & 25.9 & 25.1 & 20.9 & 27.4 & 26.1 & 26.4 & \underline{28.1} & \textbf{30.9} \\
            & LongBench & 55.1 & 56.2 & 55.4 & 52.3 & 58.7 & 57.0 & 57.5 & \underline{58.2} & \textbf{67.9} \\
        \bottomrule
    \end{tabular}
\end{table}

%% file: sections/5_conclusion.tex
\section{Conclusions and Future Work}
\label{sec:conclusion}

We present \our, a general-purpose reasoning framework that leverages Markovian transitions to minimize historical dependencies during inference. By alternating between decomposition and contraction, \our incrementally reduces complex queries into atomic subproblems, enabling scalable and modular reasoning across maths, code, and multi-hop QA tasks. Empirically, we show that \our not only scales gracefully with compute but also integrates flexibly into existing reasoning paradigms as a plug-in module.
Limitations and broader impacts of \our are provided in Appendix~\ref{sec:limitations} and~\ref{sec:impacts}.

While \our offers a promising path toward atomic reasoning, its current implementation operates solely at inference time. A natural extension is to align this structure with training-time objectives—teaching models to internalize Markovian and atomic reasoning patterns directly. This could involve supervised fine-tuning with synthetic traces, reinforcement learning over decomposition trajectories, or pretraining on datasets that promote context-isolated reasoning.

More broadly, this work lays the foundation for reasoning systems that emphasize minimal context, compositionality, and structural modularity. We hope \our serves as a stepping stone toward more efficient, interpretable, and robust reasoning with large language models.

%% file: sections/6_appendix.tex
\appendix

\section*{Appendix Overview}
This appendix is organized into three main parts: Section~\ref{sec:implementation_details_appendix} provides comprehensive implementation details including prompts, hyperparameters, and baseline configurations; Section~\ref{sec:empirical_analysis_appendix} presents detailed empirical analyses validating our framework's effectiveness; and Sections~\ref{sec:limitations}--\ref{sec:impacts} discuss limitations and broader impacts of this work.


\section{Implementation Details}
\label{sec:implementation_details_appendix}

This section provides comprehensive implementation details necessary for reproducing our experiments, including prompt templates, hyperparameter settings, and baseline method configurations.

\subsection{Prompt Templates}
\label{sec:prompt}

We present the core prompt structures used in \our for different task domains. Our framework employs four key prompt types: (1) \texttt{direct} for solving problems, (2) \texttt{decompose} for extracting DAG structures, (3) \texttt{contract} for generating simplified questions, and (4) \texttt{judge} for LLM-as-a-judge evaluation. Below we detail domain-specific implementations for mathematical reasoning, code generation, and multi-hop question answering.

\paragraph{Design Rationale.} The Multi-hop QA prompts use JSON for structured responses, while Math and Code tasks use HTML-like tags (e.g., \texttt{<answer></answer>}). This design choice reflects task-specific requirements: JSON naturally accommodates Multi-hop QA's need for structured outputs including reasoning chains and supporting evidence, while HTML tags provide clear answer demarcation for Math and Code tasks. Function parameters also vary by domain—Multi-hop QA requires context passages, Code generation needs test cases and dependency information, while Math tasks only require the question. These variations align with the inherent characteristics of each problem type rather than representing arbitrary design choices.

\subsubsection{Mathematical Reasoning}
\begin{lstlisting}[style=PythonStyle, caption={Math}]
def direct(question: str):
    instruction = """
        You are a precise math question solver. Solve the given math question step by step using a standard algebraic approach:

        QUESTION: {question}
        
        You can freely reason in your response, but please enclose the final answer within <answer></answer> tags (pure number without units and explanations)
    """
    prompt = instruction.format(question=question)
    return prompt

def decompose():
    instruction = """
        Decompose the previous reasoning trajectory into a series of sub-questions or thoughts.

        Instructions:
        1. Each sub-question or thought should list its other sub-questions or thoughts' indexes it depends (0-based, can be an empty list)
        2. Dependencies are defined as information needed in sub-question or thought that:
           - Does NOT come directly from the original question
           - MUST come from previous sub-questions or thoughts
    """
    return instruction

def contract():
    instruction = """
        Generate a simplified intermediate form of the original question based on the previous sub-questions or thoughts step by step.
        
        The previous sub-questions or thoughts with marked dependencies actually form a directed acyclic graph (DAG), where nodes whose dependencies is empty list can be regarded as independent sub-questions or thoughts.
        
        The simplified question must be:
        1. self-contained: The simplified question's description must contain all information needed to solve itself, without requiring additional information from the original question or reasoning trajectory
        2. test-time reduced: The simplified question must require fewer reasoning steps compared to the original question (these steps are reduced because these solved independent sub-problems or thoughts become known conditions in the simplified question or excluded as incorrect explorations)
        
    """
    formatter = "Last step, enclose the question within <question></question> tags"
    instruction += formatter
    return instruction

def judge(question: str, solutions: list):
    instruction = """
        Here is the original problem:
        {question}

        Here are some reference solutions:
        {solutions}
        
        Ensemble the best answer to the original problem from the solutions step by step:
    """
    formatter = "Last step, enclose the answer within <answer></answer> tags (must be an integer or decimal number without units and explanations)"
    instruction += formatter
    
    solutions_str = ""
    for i, solution in enumerate(solutions):
        solutions_str += f"solution {i}: {solution}\n"
    prompt = instruction.format(question=question, solutions=solutions_str)
    return prompt
\end{lstlisting}

\subsubsection{Code Generation}
\begin{lstlisting}[style=PythonStyle, caption={Code}]
def direct(question: str, contexts: str):
    instruction = """
        Solve the following problem step by step:
        {question}
        Your code should be a python function with format: {contexts}
        
        Please extend your reasoning process as much as possible; the longer the chain of thought, the better.

    """
    formatter = "Last step, enclose your code within ```python and ```"
    instruction += formatter
    prompt = instruction.format(question=question, contexts=contexts)
    return prompt

def decompose():
    instruction = """
        Decompose the previous reasoning trajectory into a series of sub-questions or thoughts.

        Instructions:
        1. Each sub-question or thought should list its other sub-questions or thoughts' indexes it depends (0-based, can be an empty list)
        2. Dependencies are defined as information needed in sub-question or thought that:
           - Does NOT come directly from the original question
           - MUST come from previous sub-questions or thoughts
    """
    return instruction

def contract(dag, test_cases):
    instruction = """
        Generate a simplified intermediate form of the original problem based on the variable dependency analysis.
        
        You ast.arg given a directed acyclic graph (DAG) representing the dependencies between variables in the original code:
        {dag}
        
        And the original test cases:
        {test_cases}
        
        The simplified problem must be:
        1. Self-contained: The description must contain all information needed to solve itself, without requiring additional context from the original problem
        2. Test-time reduced: The simplified problem must require fewer reasoning steps by using intermediate variables from the original code as direct inputs
        
        Your task is to:
        1. Create a simplified version of the problem that starts with intermediate variables as inputs
        2. Generate new test cases that use these intermediate variables as parameters while maintaining the exact same expected outputs as in the original test cases
        
        Do not use any code examples in your simplified problem formulation.
    """
    formatter = r"Enclose the simplified problem within <question></question> tag and the new test cases (assert codes, use \n to split each case) within <test></test> tag"
    instruction += formatter
    prompt = instruction.format(dag=dag, test_cases=test_cases)
    return prompt

def judge(question: str, solutions: list):
    instruction = """
        Here is the original problem:
        {question}

        Here are some reference solutions:
        {solutions}
        
        Give the index of the best solution as your answer.
    """
    formatter = "Last step, enclose the answer within <answer></answer> tags (0-based)"
    instruction += formatter
    
    solutions_str = ""
    for i, solution in enumerate(solutions):
        solutions_str += f"solution {i}: {solution}\n"
    prompt = instruction.format(question=question, solutions=solutions_str)
    return prompt
\end{lstlisting}

\subsubsection{Multi-hop Question Answering}
\begin{lstlisting}[style=PythonStyle, caption={Multi-hop QA}]
def direct(question: str, contexts: str):
    instruction = """
        Solve the following multi-hop question step by step:
        {question}

        CONTEXTS: 
        {contexts}

        Firstly, you need to extract the relevant supporting sentences from the original text, then cut out the continuous segments as the answer.
    """
        formatter = """
    Provide your response in this JSON format:
    {{
        "question": {question},
        "thought": "give your step by step thought process here",
        "supporting_sentences": [
            "Include ALL sentences needed to justify your answer",
            "Use ... for long sentences when appropriate"
        ],
        "answer": "Your precise answer following the instructions above" or "none" if no answer can be found
    }}
    """
    instruction += formatter
    prompt = instruction.format(question=question, contexts=contexts)
    return prompt

def decompose(question: str, trajectory: str, answer: str):
    instruction = """
        You are tasked with breaking down a multiple choice question reasoning process into sub-questions.

        Original Question: {question}
        Complete Reasoning Process: {trajectory}

        Instructions:
        1. Break down the reasoning process into a series of sub-questions
        2. Each sub-question should:
           - Be written in interrogative form
           - Have a clear answer
           - List its other sub-questions' indexes it depends (0-based, can be an empty list)
        3. Dependencies are defined as information needed to answer the current sub-question that:
           - Does NOT come directly from the original question
           - MUST come from the answers of previous sub-questions
    """
    formatter = """
        Format your response as the following JSON object:
        {{
            "thought": "<the thought process of how to step by step propose the sub-questions until the answer of the original question in the given reasoning process is obtained>",
            "sub-questions": [
                {{
                    "description": "<the description of the sub-question>", 
                    "answer": <the answer to the sub-question>,
                    "depend": [<indices of the dependent sub-questions>, ...]
                }}
            ],
            "answer": "{answer}"
        }}
    """
    return (instruction + formatter).format(question=question, trajectory=trajectory, answer=answer)

def contract(question: str, decompose_result: dict, independent: list, dependent: list):
    instruction = """
        You are a multiple choice question solver specializing in optimizing step-by-step reasoning processes. Your task is to optimize the existing reasoning trajectory into a more efficient, single self-contained question.
        
        For the original question: {question}
        
        Here are step-by-step reasoning process:
        {response}
        
        {sub_questions}
        
        Here are explanations of key concepts:
        1. self-contained: The optimized question must be solvable independently, without relying on any external information
        2. efficient: The optimized question must be simpler than the original, requiring fewer reasoning steps and having a clearer reasoning process (these steps are reduced because some solved sub-problems become known conditions in the optimized question or are excluded as incorrect explorations)
        
        Note: Since this is a multiple choice question, the optimized question must completely retain the options of the original question.
        
        You can freely reason in your response, but please enclose the your optimized question within <question></question> tags
    """
    sub_questions = """
        The following sub-questions and their answers can serve as known conditions:
        {independent}

        The descriptions of the following questions can be used to form the description of the optimized problem:
        {dependent}
        
        """
    answer = decompose_result["answer"]
    for sub_q in independent:
        sub_q.pop("depend", None)
    for sub_q in dependent:
        sub_q.pop("depend", None)
        
    sub_questions = sub_questions.format(independent=independent, dependent=dependent)
    return instruction.format(question=question, answer=answer, response=decompose_result["response"], sub_questions=sub_questions)

def judge(question: str, solutions: list):
    instruction = """
        You are a precise multiple choice question solver. Compare then synthesize the best answer from multiple solutions to select the most correct option:

        QUESTION: {question}

        SOLUTIONS:
        {solutions}
        
        Extend your chain of thought as much as possible; the longer the chain of thought, the better.

        You can freely reason in your response, even propose new reasoning to get a better answer than all solutions, but please mark the final option with <answer>single letter of your chosen option</answer> tags
    """
    
    solutions_str = ""
    for i, solution in enumerate(solutions):
        solutions_str += f"solution {i}: {solution}\n"
    prompt = instruction.format(question=question, solutions=solutions_str)
    return prompt
\end{lstlisting}

\subsection{Hyperparameter Configuration}
\label{sec:hyperparams}

\paragraph{Maximum Transition Count.}
The maximum number of transitions in the Markovian reasoning chain is a key hyperparameter that controls the depth of reasoning exploration. Theoretically, longer chains enable deeper reasoning, but practical considerations require balancing performance gains with computational efficiency. Throughout our experiments, we uniformly set the maximum transition count to 3, which empirically provides an effective trade-off (see Section~\ref{appendix:analysis} for empirical justification based on structural depth analysis).

\paragraph{Adaptive Setting.}
For query-specific optimization, the maximum transition count can be dynamically determined by analyzing the initial DAG structure. Since each transition ideally eliminates one layer of independent nodes (those without incoming edges), the depth of the initially decomposed DAG $\mathcal{G}_0$ serves as a reasonable upper bound estimate for the required number of transitions. This can be computed via a simple graph traversal without additional LLM invocations.

\paragraph{Other Hyperparameters.}
We use temperature $T=1.0$ for all LLM sampling operations to balance exploration and determinism. For integration experiments with tree-based methods (Section 4.4), we use 3 branches for ToT and vary the number of trees in FoT as \{2, 4, 8\} to study scaling behavior.

\subsection{Baseline Implementation Details}
\label{appendix:baseline_implementation}

This subsection describes our implementation of baseline methods to ensure fair and reproducible comparisons.

\subsubsection{Forest of Thoughts (FoT)}
In our implementation, we utilize the classical Tree of Thoughts (ToT) approach as the fundamental tree structure within the Forest of Thoughts framework, while maintaining several critical mechanisms from the original FoT design, including majority voting for aggregating results across different trees and expert evaluation for assessing solution quality. 

However, our implementation differs from the original FoT in certain aspects to accommodate a broader range of question types. Specifically, we remove the early stopping criteria that terminate tree splitting when nodes cannot produce valid outputs. While this mechanism is particularly effective for constrained tasks like Game-of-24 where rule-based validation is straightforward, it is less applicable to our diverse evaluation scenarios where output validity is less clearly defined. Instead, we maintain tree expansion regardless of intermediate output quality, allowing the framework to explore potentially valuable paths that might initially appear suboptimal. Additionally, we omit the Input Data Augmentation technique, as analogical reasoning approaches do not demonstrate consistent effectiveness across different question domains in our experiments.

These modifications preserve the core strengths of FoT while enhancing its adaptability to a wider range of reasoning tasks. Our implementation successfully reproduces the scaling curves reported in the original FoT paper and achieves superior performance across multiple benchmarks.

\subsubsection{AFlow}
For AFlow, we adopt the optimal workflows identified in the original work for each benchmark dataset while making necessary adaptations to our experimental setup. For mathematical reasoning tasks on MATH and GSM8K, we directly employ AFlow's proven optimal workflows. For multi-hop reasoning scenarios in LongBench, we use the workflow initially optimized for HotpotQA, as both datasets share core multi-hop reasoning characteristics. This approach ensures we leverage AFlow's strengths while maintaining consistency across similar problem types.

\subsubsection{Dataset-Specific Details}
For the MATH dataset, we filter out questions with non-integer or non-decimal answers to ensure consistent evaluation. We evaluate the first 1,000 cases from MATH for efficiency, while assessing the remaining benchmarks in their entirety.


\section{Empirical Analysis and Validation}
\label{sec:empirical_analysis_appendix}

This section presents detailed empirical analyses that validate the effectiveness of our framework, including quality metrics for DAG generation, concrete examples of the decomposition-contraction process, and statistical analyses of structural properties.

\subsection{DAG Generation Quality Assessment}
\label{appendix:dag_quality}

To evaluate the quality of our two-phase transition mechanism (decomposition and contraction), we provide comprehensive quality metrics across multiple datasets. Table~\ref{tab:dag_quality_appendix} presents three key metrics that assess different aspects of the DAG generation and state transition process.

\begin{table}[h]
\centering
\caption{DAG Generation Quality Metrics Across Benchmarks}
\label{tab:dag_quality_appendix}
\begin{tabular}{lcccc}
\toprule
\textbf{Metric} & \textbf{MATH} & \textbf{GSM8K} & \textbf{MBPP} & \textbf{LongBench} \\
\midrule
Answer Equivalence Maintenance & 99.2\% & 99.5\% & 99.7\% & 99.3\% \\
Test-time Complexity Reduction & 76.4\% & 82.1\% & 74.8\% & 79.2\% \\
LLM-as-a-Judge Selection Rate & 92.5\% & 95.8\% & 83.1\% & 91.5\% \\
\bottomrule
\end{tabular}
\end{table}

\paragraph{Evaluation Methodology.} 
Both answer equivalence and test-time complexity reduction are assessed through LLM evaluation, where the evaluator LLM is provided with $Q_i$ and $Q_{i+1}$ along with their execution processes. The LLM judges answer equivalence by examining whether the reasoning trajectory's derivation goals remain consistent, and assesses complexity reduction by analyzing the trajectory length and required reasoning steps.

\paragraph{Metric Definitions.}
\begin{itemize}
\item \textbf{Answer Equivalence Maintenance}: The probability that the contracted question $Q_{i+1}$ maintains answer equivalence with the original question $Q_0$. The consistently high rates (>99\% across all datasets) demonstrate the reliability of our transition mechanism.

\item \textbf{Test-time Complexity Reduction}: The probability that $Q_{i+1}$ successfully reduces reasoning complexity compared to $Q_i$. Rates between 74\%-82\% indicate that most transitions achieve meaningful simplification.

\item \textbf{LLM-as-a-Judge Selection Rate}: The probability that $Q_{i+1}$ is selected (or at least not rejected) by the LLM-as-a-judge when choosing among $\{\text{solve}(Q_i), \text{solve}(\mathcal{G}_i), \text{solve}(Q_{i+1})\}$. High selection rates (83\%-96\%) confirm that the generated states provide valuable reasoning improvements.
\end{itemize}

These metrics collectively demonstrate that our DAG generation and contraction process reliably maintains answer equivalence while achieving meaningful complexity reduction, validating the effectiveness of our two-phase transition design.

\subsection{Detailed DAG Decomposition and Contraction Examples}
\label{appendix:detailed_examples}

To provide concrete understanding of how the DAG decomposition and contraction mechanism works in practice, we present detailed step-by-step examples demonstrating the complete two-phase transition process.

\subsubsection{Mathematical Reasoning Example}

\paragraph{Original Problem ($Q_0$):}
\textit{There are two possible triangles ABC satisfying AB = 10, AC = b > 10, and sin B = 3/5. Find the positive difference between the lengths of side BC.}

\paragraph{Step 1: Direct Reasoning Trajectory.}
When solving $Q_0$ directly, the LLM generates the following reasoning chain:
\begin{enumerate}
\item Given sin B = 3/5, we can find cos B using the Pythagorean identity: $\cos^2 B + \sin^2 B = 1$
\item This gives us $\cos^2 B = 1 - (3/5)^2 = 1 - 9/25 = 16/25$
\item Therefore, cos B = ±4/5 (two possible values)
\item Using the Law of Cosines: $b^2 = a^2 + c^2 - 2ac \cos B$, where $a = BC$, $c = AB = 10$
\item For cos B = 4/5: $b^2 = a^2 + 100 - 20a(4/5) = a^2 - 16a + 100$
\item For cos B = -4/5: $b^2 = a^2 + 100 - 20a(-4/5) = a^2 + 16a + 100$
\item Solving these two cases yields two possible values for BC
\item The positive difference is computed from these two values
\end{enumerate}

\paragraph{Step 2: DAG Decomposition ($Q_0 \rightarrow \mathcal{G}_0$).}
The LLM decomposes this reasoning trajectory into a dependency structure:

\textbf{Node 0:} ``Calculate cos B from sin B = 3/5 using the Pythagorean identity''
\begin{itemize}
\item Dependencies: []  (no dependencies, independent subproblem)
\item Result: cos B = ±4/5
\end{itemize}

\textbf{Node 1:} ``Given AB = 10, AC = b > 10, and cos B = ±4/5, apply the Law of Cosines to find the two possible values of BC''
\begin{itemize}
\item Dependencies: [0]  (depends on the result of Node 0)
\end{itemize}

\textbf{Node 2:} ``Calculate the positive difference between the two values of BC''
\begin{itemize}
\item Dependencies: [1]  (depends on the result of Node 1)
\end{itemize}

The DAG structure is: Node 0 $\rightarrow$ Node 1 $\rightarrow$ Node 2, forming a linear chain of depth 3.

\paragraph{Step 3: Contraction ($\mathcal{G}_0 \rightarrow Q_1$).}
Nodes without incoming edges (Node 0) represent independent subproblems that can be directly solved. After solving Node 0, we obtain cos B = ±4/5. This information is incorporated into the problem statement, and nodes depending on it are reformulated:

\textbf{Contracted Question ($Q_1$):}
\textit{Given that cos B can be either 4/5 or -4/5, with AB = 10 and AC = b > 10, use the Law of Cosines to find the two possible values of BC, then calculate their positive difference.}

\textbf{Key observations:}
\begin{itemize}
\item $Q_1$ is self-contained: All necessary information (cos B values) is now explicitly stated
\item $Q_1$ maintains answer equivalence with $Q_0$: Solving $Q_1$ yields the same final answer
\item $Q_1$ has reduced test-time complexity: The trigonometric calculation is eliminated, reducing reasoning steps from 8 to approximately 5
\item The DAG depth is reduced from 3 to 2 (only Nodes 1 and 2 remain)
\end{itemize}

\paragraph{Step 4: LLM-as-a-Judge Selection.}
After generating the triplet $\{\text{solve}(Q_0), \text{solve}(\mathcal{G}_0), \text{solve}(Q_1)\}$, the LLM-as-a-judge evaluates which provides the best answer to the original problem $Q_0$. In this case:
\begin{itemize}
\item $\text{solve}(Q_0)$: Direct solution with full reasoning chain
\item $\text{solve}(\mathcal{G}_0)$: Solution by explicitly solving each node in the DAG
\item $\text{solve}(Q_1)$: Solution of the contracted problem
\end{itemize}

If $Q_1$ maintains answer equivalence (which it does), $\text{solve}(Q_1)$ will provide a valid answer and is likely to be selected due to its cleaner reasoning structure. If the contraction process had failed to maintain equivalence, $\text{solve}(Q_1)$ would produce an incorrect or nonsensical answer, and the judge would select one of the other options, naturally filtering out the failed transition.

\paragraph{Iteration Potential.}
If we continue from $Q_1$, a second transition could further decompose and contract the problem, potentially separating the two Law of Cosines calculations from the difference computation. This iterative process continues until reaching an atomic state where no further meaningful decomposition is possible.

\subsubsection{Key Insights from the Example}

This example illustrates several important aspects of our framework:

\begin{enumerate}
\item \textbf{Structural Guidance:} The DAG explicitly captures dependencies, allowing the contraction phase to identify which information can be ``baked into'' the problem statement (Node 0's result) versus which must remain as reasoning steps (Nodes 1-2).

\item \textbf{Answer Equivalence:} The contracted question $Q_1$ asks for exactly the same final answer as $Q_0$, ensuring the Markov property holds while making meaningful progress.

\item \textbf{Complexity Reduction:} By solving independent subproblems and incorporating their results, $Q_1$ requires fewer reasoning steps, reducing the test-time computational burden.

\item \textbf{Implicit Quality Control:} The LLM-as-a-judge mechanism naturally filters failed transitions—if contraction produces an invalid or non-equivalent question, it won't be selected, preventing error propagation.
\end{enumerate}

\subsection{Analysis of Structural Diversity}
\label{appendix:analysis}

To understand the structural characteristics of problems decomposed by our framework and provide empirical justification for our hyperparameter choices, we analyze the DAG structures generated from the first 1,000 questions of the MATH dataset.

\subsubsection{Graph Structure and Chain Length}
\begin{figure}[ht]
    \centering
    \includegraphics[width=0.75\textwidth]{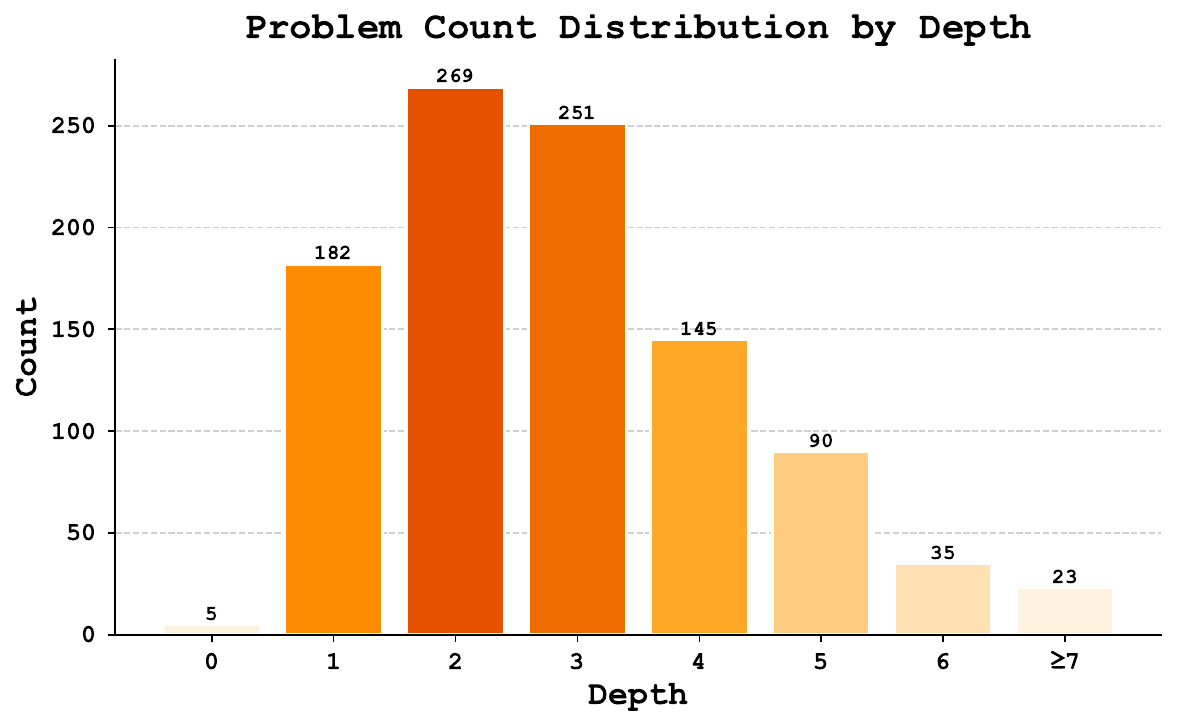} 
    \caption{Distribution of solution depths across questions. Darker orange bars indicate depths that appear more frequently in the dataset.}
    \label{fig:depth_distrib}
\end{figure}
\begin{figure}[ht]
    \centering
    \includegraphics[width=0.75\textwidth]{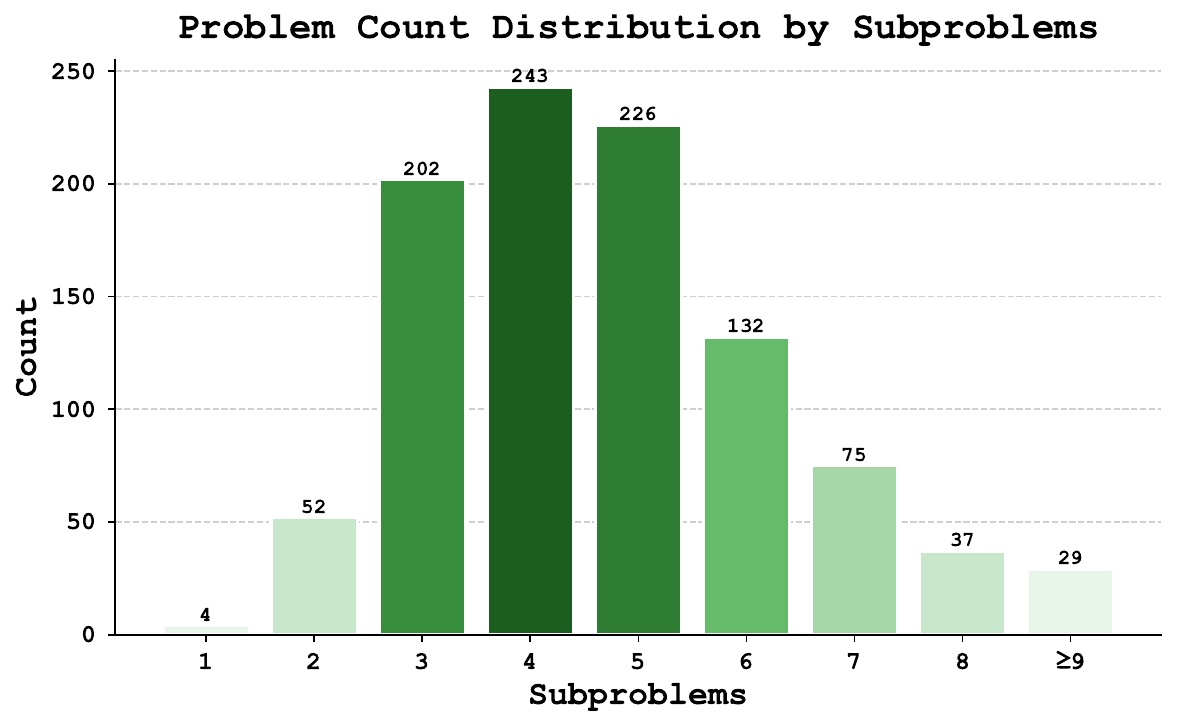} 
    \caption{Distribution of subquestion counts across questions. Darker green bars represent more common subquestion counts in the solutions.}
    \label{fig:number_distrib}
\end{figure}

Figures \ref{fig:depth_distrib} and \ref{fig:number_distrib} reveal clear structural patterns in the decomposed questions. The depth distribution (Figure \ref{fig:depth_distrib}) shows that most questions exhibit depths between 2 and 4, with depth 3 being the most common pattern. This observation provides empirical justification for our choice of maximum transition count (3) in the main experiments—the structural depth naturally aligns with the transition requirements for most problems.

Similarly, the subquestion count distribution (Figure \ref{fig:number_distrib}) indicates that questions typically decompose into 2 to 5 subquestions, with 3-4 subquestions representing the most frequent pattern. These statistics suggest that most reasoning problems naturally decompose into a small number of manageable subproblems, supporting our framework's design assumption that complex reasoning can be effectively simplified through structured decomposition.

\subsubsection{Correlation Between Structural Complexity and Performance}
\begin{figure}[ht]
   \centering
   \begin{subfigure}[b]{0.48\textwidth}
       \centering
       \includegraphics[width=\textwidth]{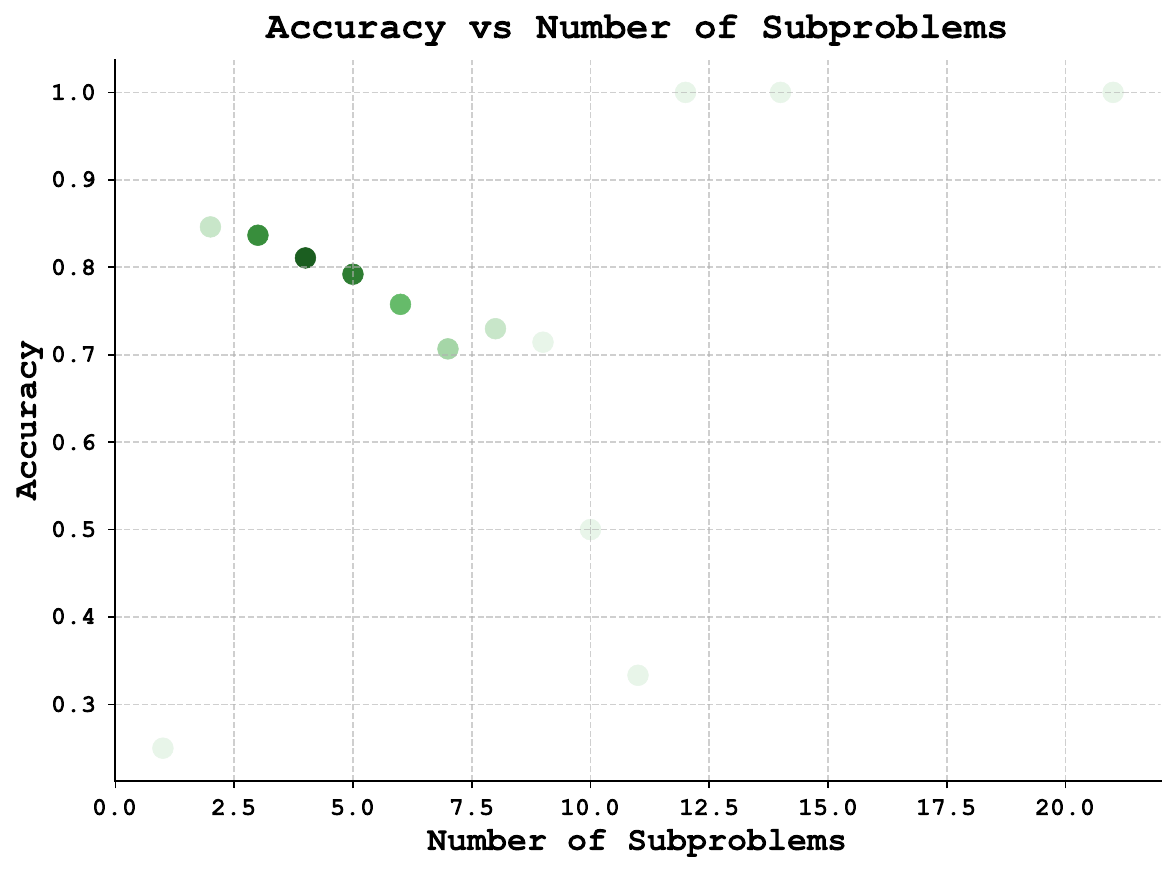}
       \caption{Number of subquestions vs accuracy}
       \label{fig:accuracy_subquestions}
   \end{subfigure}
   \hfill
   \begin{subfigure}[b]{0.48\textwidth}
       \centering
       \includegraphics[width=\textwidth]{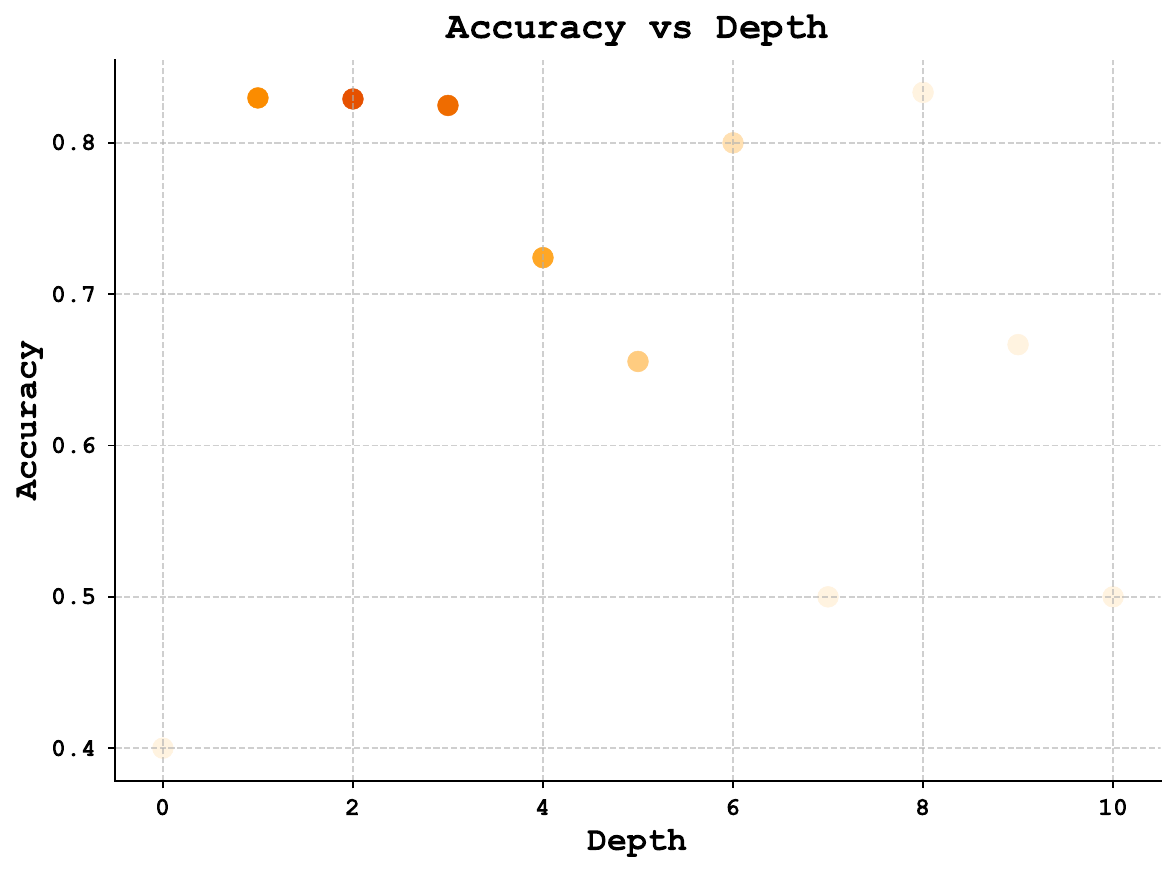}
       \caption{Solution depth vs accuracy}
       \label{fig:accuracy_depth}
   \end{subfigure}
   \caption{Correlation between structural complexity and performance. Color intensity reflects data density - darker points represent more frequent patterns.}
   \label{fig:structural_complexity}
\end{figure}

Notably, we observed correlations between these structural metrics and solution accuracy. The scatter plots reveal two important patterns: First, as shown in Figure \ref{fig:accuracy_depth}, as the depth of the solution graph increases, there is a general trend of decreasing accuracy. Second, as illustrated in Figure \ref{fig:accuracy_subquestions}, questions with more subquestions tend to show lower accuracy rates. The color intensity of the points provides additional insight - darker points represent more common structural patterns in our dataset, showing that most of our high-accuracy solutions come from questions with moderate depth and subquestion counts. This suggests that more complex question structures, characterized by either greater depth or more subquestions, pose greater challenges for question-solving systems. The decline in accuracy could be attributed to error propagation through longer solution chains and the increased cognitive load required to maintain consistency across more complex question structures.


\section{Limitations}
\label{sec:limitations}
While \our demonstrates promising results across multiple domains, several limitations remain. First, the current implementation relies on a fixed maximum transition count (set to 3), which may not be optimal for all problem types. Although we propose an adaptive setting based on initial DAG depth, a fully dynamic termination criterion would be more robust. Second, the decomposition process adds computational overhead compared to direct inference, which might be a trade-off for real-time applications. Finally, the quality of decomposition depends heavily on the underlying LLM's capability; weaker models may struggle to generate valid dependency graphs, potentially degrading performance.

\section{Broader Impacts}
\label{sec:impacts}
This work contributes to the advancement of reasoning capabilities in large language models, which has potential positive impacts in scientific discovery, education, and complex problem-solving. By making reasoning processes more structured and interpretable through decomposition, \our could help in building more trustworthy AI systems. However, as with any advanced AI capability, there are risks of misuse, such as generating more sophisticated disinformation or automating malicious code generation. We believe that the modular nature of our framework also offers opportunities for better monitoring and control, as individual reasoning steps can be inspected and verified. Future work should focus on developing robust safety guardrails that can operate within this structured reasoning framework.

%% file: aot.bbl
\begin{thebibliography}{63}
\providecommand{\natexlab}[1]{#1}
\providecommand{\url}[1]{\texttt{#1}}
\expandafter\ifx\csname urlstyle\endcsname\relax
  \providecommand{\doi}[1]{doi: #1}\else
  \providecommand{\doi}{doi: \begingroup \urlstyle{rm}\Url}\fi

\bibitem[Austin et~al.(2021)Austin, Odena, Nye, Bosma, Michalewski, Dohan, Jiang, Cai, Terry, Le, et~al.]{mbpp2021Austin}
Jacob Austin, Augustus Odena, Maxwell Nye, Maarten Bosma, Henryk Michalewski, David Dohan, Ellen Jiang, Carrie Cai, Michael Terry, Quoc Le, et~al.
\newblock Program synthesis with large language models.
\newblock \emph{arXiv preprint arXiv:2108.07732}, 2021.

\bibitem[Bai et~al.(2024)Bai, Lv, Zhang, Lyu, Tang, Huang, Du, Liu, Zeng, Hou, Dong, Tang, and Li]{longbench2024}
Yushi Bai, Xin Lv, Jiajie Zhang, Hongchang Lyu, Jiankai Tang, Zhidian Huang, Zhengxiao Du, Xiao Liu, Aohan Zeng, Lei Hou, Yuxiao Dong, Jie Tang, and Juanzi Li.
\newblock Longbench: {A} bilingual, multitask benchmark for long context understanding.
\newblock In \emph{{ACL} {(1)}}, pages 3119--3137. Association for Computational Linguistics, 2024.

\bibitem[Bechtel et~al.(2001)Bechtel, Abrahamsen, and Graham]{Smelser2001Cognitive}
W.~Bechtel, A.~Abrahamsen, and G.~Graham.
\newblock Cognitive science: History.
\newblock In Neil~J. Smelser and Paul~B. Baltes, editors, \emph{International Encyclopedia of the Social \& Behavioral Sciences}, pages 2154--2158. Pergamon, Oxford, 2001.
\newblock ISBN 978-0-08-043076-8.
\newblock \doi{https://doi.org/10.1016/B0-08-043076-7/01442-X}.
\newblock URL \url{https://www.sciencedirect.com/science/article/pii/B008043076701442X}.

\bibitem[Besta et~al.(2024)Besta, Blach, Kubicek, Gerstenberger, Podstawski, Gianinazzi, Gajda, Lehmann, Niewiadomski, Nyczyk, and Hoefler]{Besta2024got}
Maciej Besta, Nils Blach, Ales Kubicek, Robert Gerstenberger, Michal Podstawski, Lukas Gianinazzi, Joanna Gajda, Tomasz Lehmann, Hubert Niewiadomski, Piotr Nyczyk, and Torsten Hoefler.
\newblock Graph of thoughts: Solving elaborate problems with large language models.
\newblock In \emph{{AAAI}}, pages 17682--17690. {AAAI} Press, 2024.

\bibitem[Bi et~al.(2024)Bi, Han, Liu, Tang, and Wang]{Bi2024fot}
Zhenni Bi, Kai Han, Chuanjian Liu, Yehui Tang, and Yunhe Wang.
\newblock Forest-of-thought: Scaling test-time compute for enhancing {LLM} reasoning.
\newblock \emph{CoRR}, abs/2412.09078, 2024.

\bibitem[Chang et~al.(2025)Chang, Tong, Niu, Neubig, and Yue]{yeo2025demystifying}
Edward~Y. Chang, Yuxuan Tong, Morry Niu, Graham Neubig, and Xiang Yue.
\newblock Demystifying long chain-of-thought reasoning in llms.
\newblock \emph{CoRR}, abs/2502.03373, 2025.

\bibitem[Chen et~al.(2024)Chen, Davis, Hanin, Bailis, Stoica, Zaharia, and Zou]{Chen2024more}
Lingjiao Chen, Jared~Quincy Davis, Boris Hanin, Peter Bailis, Ion Stoica, Matei Zaharia, and James Zou.
\newblock Are more {LLM} calls all you need? towards scaling laws of compound inference systems.
\newblock \emph{CoRR}, abs/2403.02419, 2024.

\bibitem[Cobbe et~al.(2021)Cobbe, Kosaraju, Bavarian, Chen, Jun, Kaiser, Plappert, Tworek, Hilton, Nakano, Hesse, and Schulman]{gsm8k2021}
Karl Cobbe, Vineet Kosaraju, Mohammad Bavarian, Mark Chen, Heewoo Jun, Lukasz Kaiser, Matthias Plappert, Jerry Tworek, Jacob Hilton, Reiichiro Nakano, Christopher Hesse, and John Schulman.
\newblock Training verifiers to solve math word problems.
\newblock \emph{CoRR}, abs/2110.14168, 2021.

\bibitem[DeepSeek-AI(2025)]{deepseekR1Model}
DeepSeek-AI.
\newblock Deepseek-r1: Incentivizing reasoning capability in llms via reinforcement learning, 2025.
\newblock URL \url{https://arxiv.org/abs/2501.12948}.

\bibitem[DeepSeek{-}AI et~al.(2024)DeepSeek{-}AI, Liu, Feng, Xue, Wang, Wu, Lu, Zhao, Deng, Zhang, Ruan, Dai, Guo, Yang, Chen, Ji, Li, Lin, Dai, Luo, Hao, Chen, Li, Zhang, Bao, Xu, Wang, Zhang, Ding, Xin, Gao, Li, Qu, Cai, Liang, Guo, Ni, Li, Wang, Chen, Chen, Yuan, Qiu, Li, Song, Dong, Hu, Gao, Guan, Huang, Yu, Wang, Zhang, Xu, Xia, Zhao, Wang, Zhang, Li, Wang, Zhang, Zhang, Tang, Li, Tian, Huang, Wang, Zhang, Wang, Zhu, Chen, Du, Chen, Jin, Ge, Zhang, Pan, Wang, Xu, Zhang, Chen, Li, Lu, Zhou, Chen, Wu, Ye, Ye, Ma, Wang, Zhou, Yu, Zhou, Pan, Wang, Yun, Pei, Sun, Xiao, and Zeng]{DeepSeekV3}
DeepSeek{-}AI, Aixin Liu, Bei Feng, Bing Xue, Bingxuan Wang, Bochao Wu, Chengda Lu, Chenggang Zhao, Chengqi Deng, Chenyu Zhang, Chong Ruan, Damai Dai, Daya Guo, Dejian Yang, Deli Chen, Dongjie Ji, Erhang Li, Fangyun Lin, Fucong Dai, Fuli Luo, Guangbo Hao, Guanting Chen, Guowei Li, H.~Zhang, Han Bao, Hanwei Xu, Haocheng Wang, Haowei Zhang, Honghui Ding, Huajian Xin, Huazuo Gao, Hui Li, Hui Qu, J.~L. Cai, Jian Liang, Jianzhong Guo, Jiaqi Ni, Jiashi Li, Jiawei Wang, Jin Chen, Jingchang Chen, Jingyang Yuan, Junjie Qiu, Junlong Li, Junxiao Song, Kai Dong, Kai Hu, Kaige Gao, Kang Guan, Kexin Huang, Kuai Yu, Lean Wang, Lecong Zhang, Lei Xu, Leyi Xia, Liang Zhao, Litong Wang, Liyue Zhang, Meng Li, Miaojun Wang, Mingchuan Zhang, Minghua Zhang, Minghui Tang, Mingming Li, Ning Tian, Panpan Huang, Peiyi Wang, Peng Zhang, Qiancheng Wang, Qihao Zhu, Qinyu Chen, Qiushi Du, R.~J. Chen, R.~L. Jin, Ruiqi Ge, Ruisong Zhang, Ruizhe Pan, Runji Wang, Runxin Xu, Ruoyu Zhang, Ruyi Chen, S.~S. Li, Shanghao Lu, Shangyan Zhou, Shanhuang Chen, Shaoqing Wu, Shengfeng Ye, Shengfeng Ye, Shirong Ma, Shiyu Wang, Shuang Zhou, Shuiping Yu, Shunfeng Zhou, Shuting Pan, T.~Wang, Tao Yun, Tian Pei, Tianyu Sun, W.~L. Xiao, and Wangding Zeng.
\newblock Deepseek-v3 technical report.
\newblock \emph{CoRR}, abs/2412.19437, 2024.
\newblock \doi{10.48550/ARXIV.2412.19437}.
\newblock URL \url{https://doi.org/10.48550/arXiv.2412.19437}.

\bibitem[Ding et~al.(2024)Ding, Zhang, Wang, Xu, Ma, Zhang, Qin, Rajmohan, Lin, and Zhang]{Ding2023xot}
Ruomeng Ding, Chaoyun Zhang, Lu~Wang, Yong Xu, Minghua Ma, Wei Zhang, Si~Qin, Saravan Rajmohan, Qingwei Lin, and Dongmei Zhang.
\newblock Everything of thoughts: Defying the law of penrose triangle for thought generation.
\newblock In \emph{{ACL} (Findings)}, pages 1638--1662. Association for Computational Linguistics, 2024.

\bibitem[Hao et~al.(2023)Hao, Gu, Ma, Hong, Wang, Wang, and Hu]{hao2023rap}
Shibo Hao, Yi~Gu, Haodi Ma, Joshua~Jiahua Hong, Zhen Wang, Daisy~Zhe Wang, and Zhiting Hu.
\newblock Reasoning with language model is planning with world model.
\newblock In \emph{{EMNLP}}, pages 8154--8173. Association for Computational Linguistics, 2023.

\bibitem[Hao et~al.(2024)Hao, Gu, Luo, Liu, Shao, Wang, Xie, Ma, Samavedhi, Gao, Wang, and Hu]{hao2024llm}
Shibo Hao, Yi~Gu, Haotian Luo, Tianyang Liu, Xiyan Shao, Xinyuan Wang, Shuhua Xie, Haodi Ma, Adithya Samavedhi, Qiyue Gao, Zhen Wang, and Zhiting Hu.
\newblock {LLM} reasoners: New evaluation, library, and analysis of step-by-step reasoning with large language models.
\newblock In \emph{First Conference on Language Modeling}, 2024.
\newblock URL \url{https://openreview.net/forum?id=b0y6fbSUG0}.

\bibitem[Hendrycks et~al.(2021)Hendrycks, Burns, Kadavath, Arora, Basart, Tang, Song, and Steinhardt]{MATH2021}
Dan Hendrycks, Collin Burns, Saurav Kadavath, Akul Arora, Steven Basart, Eric Tang, Dawn Song, and Jacob Steinhardt.
\newblock Measuring mathematical problem solving with the {MATH} dataset.
\newblock In \emph{NeurIPS Datasets and Benchmarks}, 2021.

\bibitem[Ho et~al.(2020)Ho, Nguyen, Sugawara, and Aizawa]{2WikiMultiHopQA}
Xanh Ho, Anh{-}Khoa~Duong Nguyen, Saku Sugawara, and Akiko Aizawa.
\newblock Constructing {A} multi-hop {QA} dataset for comprehensive evaluation of reasoning steps.
\newblock In \emph{{COLING}}, pages 6609--6625. International Committee on Computational Linguistics, 2020.

\bibitem[Hong et~al.(2024{\natexlab{a}})Hong, Lin, Liu, Liu, Wu, Zhang, Wei, Li, Chen, Zhang, et~al.]{Hong2024data}
Sirui Hong, Yizhang Lin, Bang Liu, Bangbang Liu, Binhao Wu, Ceyao Zhang, Chenxing Wei, Danyang Li, Jiaqi Chen, Jiayi Zhang, et~al.
\newblock Data interpreter: An llm agent for data science.
\newblock \emph{arXiv preprint arXiv:2402.18679}, 2024{\natexlab{a}}.

\bibitem[Hong et~al.(2024{\natexlab{b}})Hong, Zhuge, Chen, Zheng, Cheng, Wang, Zhang, Wang, Yau, Lin, Zhou, Ran, Xiao, Wu, and Schmidhuber]{Hong2024metagpt}
Sirui Hong, Mingchen Zhuge, Jonathan Chen, Xiawu Zheng, Yuheng Cheng, Jinlin Wang, Ceyao Zhang, Zili Wang, Steven Ka~Shing Yau, Zijuan Lin, Liyang Zhou, Chenyu Ran, Lingfeng Xiao, Chenglin Wu, and J{\"{u}}rgen Schmidhuber.
\newblock Metagpt: Meta programming for {A} multi-agent collaborative framework.
\newblock In \emph{{ICLR}}. OpenReview.net, 2024{\natexlab{b}}.

\bibitem[Hou et~al.(2025)Hou, Lv, Lu, Zhang, Li, Yao, Li, Tang, and Dong]{hou2025advancing}
Zhenyu Hou, Xin Lv, Rui Lu, Jiajie Zhang, Yujiang Li, Zijun Yao, Juanzi Li, Jie Tang, and Yuxiao Dong.
\newblock Advancing language model reasoning through reinforcement learning and inference scaling.
\newblock \emph{CoRR}, abs/2501.11651, 2025.

\bibitem[Jain et~al.(2024)Jain, Han, Gu, Li, Yan, Zhang, Wang, Solar-Lezama, Sen, and Stoica]{jainlivecodebench}
Naman Jain, King Han, Alex Gu, Wen-Ding Li, Fanjia Yan, Tianjun Zhang, Sida Wang, Armando Solar-Lezama, Koushik Sen, and Ion Stoica.
\newblock Livecodebench: Holistic and contamination free evaluation of large language models for code.
\newblock In \emph{The Thirteenth International Conference on Learning Representations}, 2024.

\bibitem[Jiang et~al.(2023)Jiang, Ren, and Lin]{jiang2023llmblender}
Dongfu Jiang, Xiang Ren, and Bill~Yuchen Lin.
\newblock Llm-blender: Ensembling large language models with pairwise ranking and generative fusion.
\newblock In \emph{{ACL} {(1)}}, pages 14165--14178. Association for Computational Linguistics, 2023.

\bibitem[Kaplan et~al.(2020)Kaplan, McCandlish, Henighan, Brown, Chess, Child, Gray, Radford, Wu, and Amodei]{Kaplan2020scaling}
Jared Kaplan, Sam McCandlish, Tom Henighan, Tom~B. Brown, Benjamin Chess, Rewon Child, Scott Gray, Alec Radford, Jeffrey Wu, and Dario Amodei.
\newblock Scaling laws for neural language models.
\newblock \emph{CoRR}, abs/2001.08361, 2020.

\bibitem[Kimi et~al.(2025)Kimi, Du, Gao, Xing, Jiang, Chen, Li, Xiao, Du, Liao, et~al.]{team2025kimi}
Kimi, Angang Du, Bofei Gao, Bowei Xing, Changjiu Jiang, Cheng Chen, Cheng Li, Chenjun Xiao, Chenzhuang Du, Chonghua Liao, et~al.
\newblock Kimi k1. 5: Scaling reinforcement learning with llms.
\newblock \emph{arXiv preprint arXiv:2501.12599}, 2025.

\bibitem[Li et~al.(2025)Li, Zhang, Zhang, Zhang, Liu, Yao, Xu, Zheng, Wang, Chen, et~al.]{li2025system}
Zhong-Zhi Li, Duzhen Zhang, Ming-Liang Zhang, Jiaxin Zhang, Zengyan Liu, Yuxuan Yao, Haotian Xu, Junhao Zheng, Pei-Jie Wang, Xiuyi Chen, et~al.
\newblock From system 1 to system 2: A survey of reasoning large language models.
\newblock \emph{arXiv preprint arXiv:2502.17419}, 2025.

\bibitem[Madaan et~al.(2023)Madaan, Tandon, Gupta, Hallinan, Gao, Wiegreffe, Alon, Dziri, Prabhumoye, Yang, Gupta, Majumder, Hermann, Welleck, Yazdanbakhsh, and Clark]{Madaan2023selfrefine}
Aman Madaan, Niket Tandon, Prakhar Gupta, Skyler Hallinan, Luyu Gao, Sarah Wiegreffe, Uri Alon, Nouha Dziri, Shrimai Prabhumoye, Yiming Yang, Shashank Gupta, Bodhisattwa~Prasad Majumder, Katherine Hermann, Sean Welleck, Amir Yazdanbakhsh, and Peter Clark.
\newblock Self-refine: Iterative refinement with self-feedback.
\newblock In \emph{NeurIPS}, 2023.

\bibitem[Muennighoff et~al.(2025)Muennighoff, Yang, Shi, Li, Fei{-}Fei, Hajishirzi, Zettlemoyer, Liang, Cand{\`{e}}s, and Hashimoto]{muennighoff2025s1}
Niklas Muennighoff, Zitong Yang, Weijia Shi, Xiang~Lisa Li, Li~Fei{-}Fei, Hannaneh Hajishirzi, Luke Zettlemoyer, Percy Liang, Emmanuel~J. Cand{\`{e}}s, and Tatsunori Hashimoto.
\newblock s1: Simple test-time scaling.
\newblock \emph{CoRR}, abs/2501.19393, 2025.

\bibitem[OpenAI(2024{\natexlab{a}})]{GPT4omini}
OpenAI.
\newblock Gpt-4o mini: advancing cost-efficient intelligence, 2024{\natexlab{a}}.
\newblock URL \url{https://openai.com/index/gpt-4o-mini-advancing-cost-efficient-intelligence/}.

\bibitem[OpenAI(2024{\natexlab{b}})]{O12024}
OpenAI.
\newblock Introducing openai o1, 2024{\natexlab{b}}.
\newblock URL \url{https://openai.com/o1/}.

\bibitem[OpenAI(2025)]{O3miniModel}
OpenAI.
\newblock {OpenAI} o3-mini: Pushing the frontier of cost-effective reasoning, 2025.
\newblock URL \url{https://openai.com/index/openai-o3-mini/}.

\bibitem[Saad{-}Falcon et~al.(2024)Saad{-}Falcon, Lafuente, Natarajan, Maru, Todorov, Guha, Buchanan, Chen, Guha, R{\'{e}}, and Mirhoseini]{Falcon2024archon}
Jon Saad{-}Falcon, Adrian~Gamarra Lafuente, Shlok Natarajan, Nahum Maru, Hristo Todorov, Etash Guha, Estefany~Kelly Buchanan, Mayee~F. Chen, Neel Guha, Christopher R{\'{e}}, and Azalia Mirhoseini.
\newblock Archon: An architecture search framework for inference-time techniques.
\newblock \emph{CoRR}, abs/2409.15254, 2024.

\bibitem[Snell et~al.(2024)Snell, Lee, Xu, and Kumar]{Snell2024ScalingLT}
Charlie Snell, Jaehoon Lee, Kelvin Xu, and Aviral Kumar.
\newblock Scaling {LLM} test-time compute optimally can be more effective than scaling model parameters.
\newblock \emph{CoRR}, abs/2408.03314, 2024.

\bibitem[Song et~al.(2025{\natexlab{a}})Song, Jiang, Min, Chen, Chen, Zhao, Fang, and Wen]{song2025r1}
Huatong Song, Jinhao Jiang, Yingqian Min, Jie Chen, Zhipeng Chen, Wayne~Xin Zhao, Lei Fang, and Ji-Rong Wen.
\newblock R1-searcher: Incentivizing the search capability in llms via reinforcement learning.
\newblock \emph{arXiv preprint arXiv:2503.05592}, 2025{\natexlab{a}}.

\bibitem[Song et~al.(2025{\natexlab{b}})Song, Jiang, Tian, Chen, Wu, Zhao, Min, Zhao, Fang, and Wen]{song2025r1pp}
Huatong Song, Jinhao Jiang, Wenqing Tian, Zhipeng Chen, Yuhuan Wu, Jiahao Zhao, Yingqian Min, Wayne~Xin Zhao, Lei Fang, and Ji-Rong Wen.
\newblock R1-searcher++: Incentivizing the dynamic knowledge acquisition of llms via reinforcement learning.
\newblock \emph{arXiv preprint arXiv:2505.17005}, 2025{\natexlab{b}}.

\bibitem[Sun et~al.(2025)Sun, Song, Wang, Ren, Jiang, Zhang, Bai, Deng, Zhao, Liu, et~al.]{sun2025simpledeepsearcher}
Shuang Sun, Huatong Song, Yuhao Wang, Ruiyang Ren, Jinhao Jiang, Junjie Zhang, Fei Bai, Jia Deng, Wayne~Xin Zhao, Zheng Liu, et~al.
\newblock Simpledeepsearcher: Deep information seeking via web-powered reasoning trajectory synthesis.
\newblock \emph{arXiv preprint arXiv:2505.16834}, 2025.

\bibitem[Trivedi et~al.(2022)Trivedi, Balasubramanian, Khot, and Sabharwal]{Musique2022}
Harsh Trivedi, Niranjan Balasubramanian, Tushar Khot, and Ashish Sabharwal.
\newblock Musique: Multihop questions via single-hop question composition.
\newblock \emph{Trans. Assoc. Comput. Linguistics}, 10:\penalty0 539--554, 2022.

\bibitem[Wang et~al.(2023{\natexlab{a}})Wang, Xu, Lan, Hu, Lan, Lee, and Lim]{wang2023planandsolve}
Lei Wang, Wanyu Xu, Yihuai Lan, Zhiqiang Hu, Yunshi Lan, Roy~Ka{-}Wei Lee, and Ee{-}Peng Lim.
\newblock Plan-and-solve prompting: Improving zero-shot chain-of-thought reasoning by large language models.
\newblock In \emph{{ACL} {(1)}}, pages 2609--2634. Association for Computational Linguistics, 2023{\natexlab{a}}.

\bibitem[Wang et~al.(2023{\natexlab{b}})Wang, Wei, Schuurmans, Le, Chi, Narang, Chowdhery, and Zhou]{Wang2023cotsc}
Xuezhi Wang, Jason Wei, Dale Schuurmans, Quoc~V. Le, Ed~H. Chi, Sharan Narang, Aakanksha Chowdhery, and Denny Zhou.
\newblock Self-consistency improves chain of thought reasoning in language models.
\newblock In \emph{{ICLR}}. OpenReview.net, 2023{\natexlab{b}}.

\bibitem[Wei et~al.(2022)Wei, Wang, Schuurmans, Bosma, Ichter, Xia, Chi, Le, and Zhou]{Wei2022cot}
Jason Wei, Xuezhi Wang, Dale Schuurmans, Maarten Bosma, Brian Ichter, Fei Xia, Ed~H. Chi, Quoc~V. Le, and Denny Zhou.
\newblock Chain-of-thought prompting elicits reasoning in large language models.
\newblock In \emph{NeurIPS}, 2022.

\bibitem[Xiang et~al.(2024)Xiang, Liu, Jiang, Nie, Huang, Fan, Li, Huang, Zeng, Han, Hong, Xu, and Liang]{Xiang2024AtomThink}
Kun Xiang, Zhili Liu, Zihao Jiang, Yunshuang Nie, Runhui Huang, Haoxiang Fan, Hanhui Li, Weiran Huang, Yihan Zeng, Jianhua Han, Lanqing Hong, Hang Xu, and Xiaodan Liang.
\newblock Atomthink: {A} slow thinking framework for multimodal mathematical reasoning.
\newblock \emph{CoRR}, abs/2411.11930, 2024.

\bibitem[Xiang et~al.(2025)Xiang, Liu, Jiang, Nie, Cai, Yin, Huang, Fan, Li, Huang, et~al.]{xiang2025can}
Kun Xiang, Zhili Liu, Zihao Jiang, Yunshuang Nie, Kaixin Cai, Yiyang Yin, Runhui Huang, Haoxiang Fan, Hanhui Li, Weiran Huang, et~al.
\newblock Can atomic step decomposition enhance the self-structured reasoning of multimodal large models?
\newblock \emph{arXiv preprint arXiv:2503.06252}, 2025.

\bibitem[Xin et~al.(2024)Xin, Liu, Yao, Lee, Cao, Hou, and Li]{Xin2024atomr}
Amy Xin, Jinxin Liu, Zijun Yao, Zhicheng Lee, Shulin Cao, Lei Hou, and Juanzi Li.
\newblock Atomr: Atomic operator-empowered large language models for heterogeneous knowledge reasoning.
\newblock \emph{CoRR}, abs/2411.16495, 2024.

\bibitem[Xu et~al.(2025)Xu, Xie, Zhao, and He]{CoD2025}
Silei Xu, Wenhao Xie, Lingxiao Zhao, and Pengcheng He.
\newblock Chain of draft: Thinking faster by writing less.
\newblock \emph{CoRR}, abs/2502.18600, 2025.
\newblock \doi{10.48550/ARXIV.2502.18600}.
\newblock URL \url{https://doi.org/10.48550/arXiv.2502.18600}.

\bibitem[Yang et~al.(2024)Yang, Fan, and Liao]{Yang2024mcot}
Wen Yang, Kai Fan, and Minpeng Liao.
\newblock Markov chain of thought for efficient mathematical reasoning.
\newblock \emph{CoRR}, abs/2410.17635, 2024.

\bibitem[Yang et~al.(2018)Yang, Qi, Zhang, Bengio, Cohen, Salakhutdinov, and Manning]{HotpotQA2018}
Zhilin Yang, Peng Qi, Saizheng Zhang, Yoshua Bengio, William~W. Cohen, Ruslan Salakhutdinov, and Christopher~D. Manning.
\newblock Hotpotqa: {A} dataset for diverse, explainable multi-hop question answering.
\newblock In \emph{{EMNLP}}, pages 2369--2380. Association for Computational Linguistics, 2018.

\bibitem[Yao et~al.(2023)Yao, Yu, Zhao, Shafran, Griffiths, Cao, and Narasimhan]{Yao2023tot}
Shunyu Yao, Dian Yu, Jeffrey Zhao, Izhak Shafran, Tom Griffiths, Yuan Cao, and Karthik Narasimhan.
\newblock Tree of thoughts: Deliberate problem solving with large language models.
\newblock In \emph{NeurIPS}, 2023.

\bibitem[Yao et~al.(2025{\natexlab{a}})Yao, Ren, Liao, and Liu]{yao2025unveiling}
Xinhao Yao, Ruifeng Ren, Yun Liao, and Yong Liu.
\newblock Unveiling the mechanisms of explicit cot training: How chain-of-thought enhances reasoning generalization.
\newblock \emph{CoRR}, abs/2502.04667, 2025{\natexlab{a}}.

\bibitem[Yao et~al.(2025{\natexlab{b}})Yao, Wu, LIU, Luo, Han, Liu, Guo, and Song]{yao2025determine}
Yuxuan Yao, Han Wu, Mingyang LIU, Sichun Luo, Xiongwei Han, Jie Liu, Zhijiang Guo, and Linqi Song.
\newblock Determine-then-ensemble: Necessity of top-k union for large language model ensembling.
\newblock In \emph{The Thirteenth International Conference on Learning Representations}, 2025{\natexlab{b}}.
\newblock URL \url{https://openreview.net/forum?id=FDnZFpHmU4}.

\bibitem[Yasunaga et~al.(2024)Yasunaga, Chen, Li, Pasupat, Leskovec, Liang, Chi, and Zhou]{Yasunaga2024AP}
Michihiro Yasunaga, Xinyun Chen, Yujia Li, Panupong Pasupat, Jure Leskovec, Percy Liang, Ed~H. Chi, and Denny Zhou.
\newblock Large language models as analogical reasoners.
\newblock In \emph{{ICLR}}. OpenReview.net, 2024.

\bibitem[Ye et~al.(2025)Ye, Huang, Xiao, Chern, Xia, and Liu]{ye2025limo}
Yixin Ye, Zhen Huang, Yang Xiao, Ethan Chern, Shijie Xia, and Pengfei Liu.
\newblock Limo: Less is more for reasoning.
\newblock \emph{arXiv preprint arXiv:2502.03387}, 2025.

\bibitem[Yu et~al.(2025)Yu, Zhang, Zhu, Yuan, Zuo, Yue, Fan, Liu, Liu, Liu, et~al.]{yu2025dapo}
Qiying Yu, Zheng Zhang, Ruofei Zhu, Yufeng Yuan, Xiaochen Zuo, Yu~Yue, Tiantian Fan, Gaohong Liu, Lingjun Liu, Xin Liu, et~al.
\newblock Dapo: An open-source llm reinforcement learning system at scale.
\newblock \emph{arXiv preprint arXiv:2503.14476}, 2025.

\bibitem[Zekri et~al.(2024)Zekri, Odonnat, Benechehab, Bleistein, Boull{\'{e}}, and Redko]{Zekri2024Large}
Oussama Zekri, Ambroise Odonnat, Abdelhakim Benechehab, Linus Bleistein, Nicolas Boull{\'{e}}, and Ievgen Redko.
\newblock Large language models as markov chains.
\newblock \emph{CoRR}, abs/2410.02724, 2024.

\bibitem[Zeng et~al.(2025)Zeng, Huang, Liu, Liu, He, Ma, and He]{zeng2025simplerl}
Weihao Zeng, Yuzhen Huang, Qian Liu, Wei Liu, Keqing He, Zejun Ma, and Junxian He.
\newblock Simplerl-zoo: Investigating and taming zero reinforcement learning for open base models in the wild.
\newblock \emph{arXiv preprint arXiv:2503.18892}, 2025.

\bibitem[Zhan et~al.(2025)Zhan, Zhao, Li, Liu, Zhang, Ai, Mao, Wang, Zhang, and Ma]{zhan2025evaluating}
Jingtao Zhan, Jiahao Zhao, Jiayu Li, Yiqun Liu, Bo~Zhang, Qingyao Ai, Jiaxin Mao, Hongning Wang, Min Zhang, and Shaoping Ma.
\newblock Evaluating intelligence via trial and error.
\newblock \emph{arXiv preprint arXiv:2502.18858}, 2025.

\bibitem[Zhang et~al.(2025)Zhang, Chen, Wan, Chang, Cheng, Wang, Hu, and Bai]{zhang2025evoflow}
Guibin Zhang, Kaijie Chen, Guancheng Wan, Heng Chang, Hong Cheng, Kun Wang, Shuyue Hu, and Lei Bai.
\newblock Evoflow: Evolving diverse agentic workflows on the fly.
\newblock \emph{CoRR}, abs/2502.07373, 2025.

\bibitem[Zhang et~al.(2024{\natexlab{a}})Zhang, Xiang, Yu, Teng, Chen, Chen, Zhuge, Cheng, Hong, Wang, et~al.]{zhang2024aflow}
Jiayi Zhang, Jinyu Xiang, Zhaoyang Yu, Fengwei Teng, Xionghui Chen, Jiaqi Chen, Mingchen Zhuge, Xin Cheng, Sirui Hong, Jinlin Wang, et~al.
\newblock Aflow: Automating agentic workflow generation.
\newblock \emph{arXiv preprint arXiv:2410.10762}, 2024{\natexlab{a}}.

\bibitem[Zhang et~al.(2024{\natexlab{b}})Zhang, Zhao, Zhao, Yu, He, and Fan]{zhang2024mobileexperts}
Jiayi Zhang, Chuang Zhao, Yihan Zhao, Zhaoyang Yu, Ming He, and Jianping Fan.
\newblock Mobileexperts: A dynamic tool-enabled agent team in mobile devices.
\newblock \emph{arXiv preprint arXiv:2407.03913}, 2024{\natexlab{b}}.

\bibitem[Zhang and Liu(2024)]{zhang2024tse}
Jinghan Zhang and Kunpeng Liu.
\newblock Thought space explorer: Navigating and expanding thought space for large language model reasoning.
\newblock In \emph{2024 IEEE International Conference on Big Data (BigData)}, pages 8259--8251. IEEE, 2024.

\bibitem[Zhang et~al.(2024{\natexlab{c}})Zhang, Yuan, and Yao]{Zhang2024dot}
Yifan Zhang, Yang Yuan, and Andrew~Chi{-}Chih Yao.
\newblock On the diagram of thought.
\newblock \emph{CoRR}, abs/2409.10038, 2024{\natexlab{c}}.

\bibitem[Zhang et~al.(2023)Zhang, Zhang, Li, and Smola]{Zhang2024autocot}
Zhuosheng Zhang, Aston Zhang, Mu~Li, and Alex Smola.
\newblock Automatic chain of thought prompting in large language models.
\newblock In \emph{{ICLR}}. OpenReview.net, 2023.

\bibitem[Zheng et~al.(2023)Zheng, Liu, Xie, Li, and Li]{zheng2023progressive}
Chuanyang Zheng, Zhengying Liu, Enze Xie, Zhenguo Li, and Yu~Li.
\newblock Progressive-hint prompting improves reasoning in large language models.
\newblock \emph{CoRR}, abs/2304.09797, 2023.

\bibitem[Zheng et~al.(2024)Zheng, Mishra, Chen, Cheng, Chi, Le, and Zhou]{zheng2024stepback}
Huaixiu~Steven Zheng, Swaroop Mishra, Xinyun Chen, Heng{-}Tze Cheng, Ed~H. Chi, Quoc~V. Le, and Denny Zhou.
\newblock Take a step back: Evoking reasoning via abstraction in large language models.
\newblock In \emph{{ICLR}}. OpenReview.net, 2024.

\bibitem[Zhou et~al.(2024{\natexlab{a}})Zhou, Yan, Shlapentokh{-}Rothman, Wang, and Wang]{Zhou2024lats}
Andy Zhou, Kai Yan, Michal Shlapentokh{-}Rothman, Haohan Wang, and Yu{-}Xiong Wang.
\newblock Language agent tree search unifies reasoning, acting, and planning in language models.
\newblock In \emph{{ICML}}. OpenReview.net, 2024{\natexlab{a}}.

\bibitem[Zhou et~al.(2023)Zhou, Sch{\"{a}}rli, Hou, Wei, Scales, Wang, Schuurmans, Cui, Bousquet, Le, and Chi]{zhou2023least}
Denny Zhou, Nathanael Sch{\"{a}}rli, Le~Hou, Jason Wei, Nathan Scales, Xuezhi Wang, Dale Schuurmans, Claire Cui, Olivier Bousquet, Quoc~V. Le, and Ed~H. Chi.
\newblock Least-to-most prompting enables complex reasoning in large language models.
\newblock In \emph{{ICLR}}. OpenReview.net, 2023.

\bibitem[Zhou et~al.(2024{\natexlab{b}})Zhou, Pujara, Ren, Chen, Cheng, Le, Chi, Zhou, Mishra, and Zheng]{Zhou2024selfdiscover}
Pei Zhou, Jay Pujara, Xiang Ren, Xinyun Chen, Heng{-}Tze Cheng, Quoc~V. Le, Ed~H. Chi, Denny Zhou, Swaroop Mishra, and Huaixiu~Steven Zheng.
\newblock {SELF-DISCOVER:} large language models self-compose reasoning structures.
\newblock In \emph{NeurIPS}, 2024{\natexlab{b}}.

\end{thebibliography}
